\documentclass[11pt,table]{article}
\usepackage[]{emnlp2021}
\usepackage{enumitem}
\usepackage{times}
\usepackage{latexsym}
\usepackage{graphicx}

\usepackage{algorithm}

\usepackage[utf8]{inputenc}
\usepackage{algorithmic}

\usepackage{microtype}
\usepackage{multirow}

\usepackage[normalem]{ulem}
\usepackage{subfigure}
\usepackage{xcolor}

\title{Multi-granularity Textual Adversarial Attack with Behavior Cloning}

 \author{Author 1 \and ... \and Author n \\
         Address line \\ ... \\ Address line}
\author{
Yangyi Chen$^{1,2}$\thanks{\ \  Work done during internship at CCIIP  } \hspace{0.3em},
Jin Su$^{1,2*}$ \thanks{\ \ Indicates equal contribution  },
Wei Wei$^{1}$\thanks{\ \  Corresponding author}\hspace{0.3em}

\\ 
$^{1}$Cognitive Computing and Intelligent Information Processing Laboratory,  School of Computer \\ Science and Technology
Huazhong University of Science and Tehchnology
 \\
$^{2}$School of Software Engineering,
Huazhong University of Science and Tehchnology\\

{ \{yangyichen6666, sujinltenjoy\}@gmail.com}
\\
{\tt weiw@hust.edu.cn}
}

\begin{document}
\maketitle

\begin{abstract}
Recently, the textual adversarial attack models become increasingly popular due to their successful in estimating the robustness of NLP models.
However, existing works have obvious deficiencies.~(1)~They usually consider only a single granularity of modification strategies (e.g. word-level or sentence-level), which is insufficient to explore the holistic textual space for generation;  
(2) They need to query victim models hundreds of times to make a successful attack, which is highly inefficient in practice. To address such problems, in this paper we propose \textbf{MAYA}, a \textbf{M}ulti-gr\textbf{A}nularit\textbf{Y} \textbf{A}ttack model to effectively generate high-quality adversarial samples with fewer queries to victim models. 
Furthermore, we propose a reinforcement-learning based method to train a multi-granularity attack agent through behavior cloning with the expert knowledge from our MAYA algorithm to further reduce the query times. Additionally, we also adapt the agent to attack black-box models that only output labels without confidence scores. We conduct comprehensive experiments to evaluate our attack models by attacking BiLSTM, BERT and RoBERTa in two different black-box attack settings and three benchmark datasets. Experimental results show that our models achieve overall better attacking performance and produce more fluent and grammatical adversarial samples compared to baseline models. Besides, our adversarial attack agent significantly reduces the query times in both attack settings. Our codes are released at~\url{https://github.com/Yangyi-Chen/MAYA}.

\end{abstract}

\section{Introduction}
Deep learning has been proven to be successful for many real-world applications such as spam filtering \citep{inproceedings}, autonomous driving \citep{chen2017multi}, and face recognition \citep{sun2015deepid3}. However, these powerful models are vulnerable to adversarial samples, crafted by adding small, human-imperceptible perturbations to the input \citep{goodfellow2014explaining, szegedy2013intriguing}. In the domain of computer vision, numerous adversarial attack models have been proposed to benchmark and interpret black-box deep learning models \citep{dong2018boosting, moosavi2016deepfool, carlini2017towards, kurakin2017adversarial} and corresponding defense methods have also been proposed to tackle adversarial security issues \citep{dziugaite2016study, xie2017mitigating,kurakin2017adversarial, tramer2017ensemble}. 

However, crafting textual adversarial samples is more challenging due to the discrete and non-differentiable nature of text space. Indeed, most existing works focus on a single granularity of modification strategies, such as sentence-level \citep{jia-liang-2017-adversarial, iyyer-etal-2018-adversarial}, word-level \citep{zang-etal-2020-word, ren-etal-2019-generating} or character-level \citep{eger-etal-2019-text}.
Thus, none of such attack models find the optimal solution through multi-granularities for launching attacks simultaneously, which is more efficient to generate high-performance and effective adversarial samples while preserving semantic consistency and language fluency. 
To this end, we propose a simple and novel attack model targeting on multiple kinds of granularity called \textbf{MAYA}, which achieves higher attack success rate with fewer queries to victim models and produces high-quality adversarial samples compared to baseline attack models. Specifically, we add perturbations to the original sentence via rewriting its constituents according to the strict grammatical constraints.

Besides, almost all current attack models need to query victim models hundreds or even thousands of times to launch a successful attack\footnote{On average, PSO attack model \citep{zang-etal-2020-word} need to query victim models about 5000 times in SST-2} and assume the victim models may output the confidence scores of their predictions, which is neither efficient nor practical in real-world situations. To alleviate such problems, we propose to train a multi-granularity attack agent called \textbf{MAYA$_\pi$} through behavior cloning \citep{ijcai2018-0687} with the expert knowledge from our MAYA algorithm. 

We conduct exhaustive experiments including attacking three victim models over three benchmark datasets in two different black-box settings, namely score-based and decision-based attack, to evaluate the effectiveness of our attack models. While the former supposes the labels and the confidence scores of the victim models are available, the latter assumes only the label information can be accessed while the other is unknown, which is more challenging and rarely investigated.

Experimental results demonstrate the superiority of our attack models. Specifically, MAYA overall outperforms all baseline models in terms of attack success rate, attack efficiency, and quality of adversarial samples. MAYA$_\pi$ achieves comparable attack success rate and adversarial samples quality with baseline models while significantly reduces the query times in two black-box settings. Furthermore, we apply MAYA$_\pi$ to attack open-source NLP frameworks to demonstrate its practicality and effectiveness in practice.


To summarize, the main contributions of this paper are as follows:

\begin{itemize} [topsep=1pt, partopsep=1pt, leftmargin=12pt, itemsep=-2pt]

\item Different from previous works that only concentrate on a single granularity, we propose an effective multi-granularity attack model to generate fluent and grammatical adversarial samples with fewer queries to victim models.
\item We propose a RL-based method to train an agent through Behavior Cloning with the expert knowledge from our multi-granularity attack model and demonstrate its efficiency and power in two black-box settings, proving the effectiveness of our adapted imitation algorithm.
\item We successfully handle the issues of decision-based black-box attack, which is rarely investigated in NLP.

\end{itemize}



\section{Related Work}
Existing textual adversarial attack models can be roughly categorized according to the granularity of modification, e.g., character-level, word-level,  sentence-level.		

Sentence-level attack models often contain paraphrasing original sentences following pre-defined syntax patterns \citep{iyyer-etal-2018-adversarial}, adding an irrelevant sentence to the end of the passage to distract models \citep{jia-liang-2017-adversarial}, and conducting domain shift on original sentences \citep{wang-etal-2020-cat}. However, sentence-level attacks usually neglect fine-grained granularity, such as word-level, resulting in low attack success rate.

Word-level attack is relatively more investigated and can be modeled as a combinatorial optimization problem \citep{zang-etal-2020-word}, including finding substitution words and searching for adversarial samples. The methods of finding candidate substitutes mainly focus on the similarity of word embeddings \citep{Jin2019IsBR}, WordNet synonyms \citep{ren-etal-2019-generating}, HowNet synonyms \citep{zang-etal-2020-word}, and Masked Language Model (MLM) \citep{li-etal-2020-bert-attack}. Generally, the search algorithms involve greedy search algorithm \citep{ren-etal-2019-generating, ijcai2018-0585,Jin2019IsBR}, genetic algorithm \citep{alzantot-etal-2018-generating}, and particle swarm optimization \citep{zang-etal-2020-word}. Although these attack models can achieve relatively high attacking performance, considering only a single granularity restricts the upper bound of word-level attack models' performance and almost all these models need to query victim models hundreds of times to launch a successful attack.

Character-level attacks make different modifications to words such as swapping, deleting, and inserting characters \citep{ebrahimi-etal-2018-adversarial, belinkov2018synthetic, 8424632}. These attack models often craft ungrammatical adversarial samples and can be easily defended \citep{pruthi-etal-2019-combating, jones-etal-2020-robust}. Hence, in this work, we do not incorporate character-level modification into our multi-granularity framework.

To sum up, all above models only consider a single granularity and thus are insufficient in exploring the textual space for generation. So, we propose to launch attacks on multiple granularities in this paper. Experimental results demonstrate the effectiveness and efficiency of our method.

\section{Methodology}
In this section, we first describe our multi-granularity attack (\textbf{MAYA}) model in detail. Then we introduce how to train an attack agent, denoted as \textbf{MAYA$_\pi$}, with the knowledge from our MAYA algorithm. Finally, we describe how we adapt MAYA$_\pi$ to perform decision-based black-box attack.


\subsection{Multi-granularity Adversarial Attack}
Our MAYA model incorporates three parts, namely generating adversarial candidates (\textbf{Generate}), verifying the successful attack (\textbf{Verify}), and picking the most potential candidate if no successful attack found (\textbf{Pick}). The whole process is shown as pseudocode in Appendix \ref{sec:appendix}.

\paragraph{Generate} Given the input sentence $S=[w_0, ... , w_i, ..., w_n]$, we first conduct constituency parsing on the original sentence using SuPar \citep{zhang-etal-2020-fast} to obtain its constituents. Then we generate adversarial candidates from two different perspectives.

First, for each constituent (including the whole sentence), i.e., each granularity of modification, except word-level, 
we employ various paraphrase models to generate adversarial samples via rewriting the specified constituents while keep the reset unchanged. 
However, such setting is solely a local modification which may cause syntactic inconsistency of the whole sentence, and thus we adopt the following rules to make the process more rational:
\begin{itemize} [topsep=1pt, partopsep=1pt, leftmargin=12pt, itemsep=-2pt]

\item The number of grammatical mistakes of the generated adversarial candidates must be less than or equal to the one of the original sentence, which can be checked by Language-Tool\footnote{\url{ https://www.languagetool.org}}.
\item The chosen adversarial candidate should be the one that is the most similar to the original one, i.e., preserving most of semantic information of the given sentence as much as possible. Specifically, Sentence-BERT \citep{reimers2019sentence} is adopted for encoding the sentence and its candidates and we consider to employ a similarity function (e.g., cosine) to measure the semantic perseverance. \end{itemize}
The filtered candidates are collected into a set (denoted as $V_p$).

Next, for word-level perturbation, we mask words in the original sentence one by one to generate corresponding adversarial candidates. Specifically, for $w_i$, we generate adversarial candidate $S_{w_i}=[w_0, ..., [MASK], ..., w_n]$. We collect all adversarial candidates generated in this way into a set (denoted as $V_s$).

\paragraph{Verify} Given all adversarial candidates, we query the victim model for decisions and confidence scores. If there doesn't exist an adversarial candidate successfully fools the victim model, we enter into the \textbf{Pick} step, which we will discuss later. If one or more than one successful adversarial candidates found, there are three different cases that we address differently. 
First, if all successful candidates come from $V_p$, we choose the one that retains the most semantics measured by cosine similarity of sentence embeddings as the final adversarial sample. 
Second, if successful candidates come from both $V_p$ and $V_s$, we only choose the candidates from $V_p$ following the same rule in the first case. The reason we ignore candidates from $V_s$ is that we need to fill the $[MASK]$ token with substitutes and continually query the victim model for decisions, which is inefficient in the case we already have successful candidates from $V_p$.
Finally, if all successful candidates come from $V_s$, we need to fill the $[MASK]$ token with substitutes to verify their success. Due to the same workflow in the \textbf{Pick} step, we directly view each successful candidate as $S^{'}$ and move to the second case in the \textbf{Pick} step. 

\begin{figure*}[ht]
\centering
\includegraphics[width=0.95\textwidth]{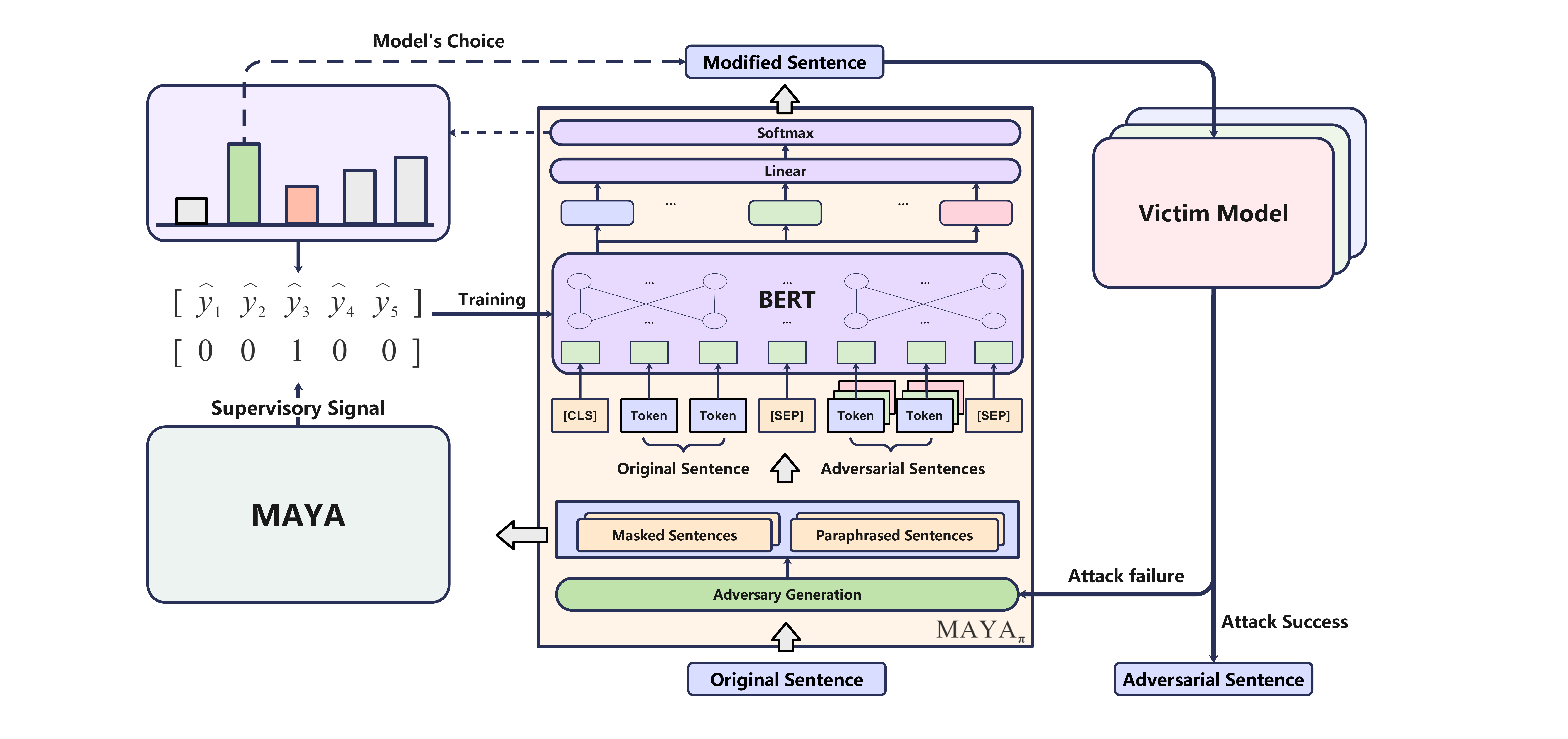}
\caption{(Left) The training process of MAYA$_\pi$. We treat the selection of candidates as a multi-class classification problem with supervisory signals from our MAYA algorithm. (Middle) The architecture and workflow of MAYA$_\pi$. (Right) The process to launch an adversarial attack against victim models. 
}
\label{fig:pathdemo}
\end{figure*}

\paragraph{Pick} If no successful candidate found, we need to pick the most potential candidate as the new sentence and repeat the same \textbf{Generate} and \textbf{Verify} procedures to find the adversarial sample. Our criterion is the decrease of the victim model's confidence score. Here we denote the candidate that causes the biggest drop in the victim model's confidence score as $S^{'}$. There are also two different cases. 
First, when $S^{'}$ comes from $V_p$, we directly choose $S^{'}$ as the most potential candidate and return to the \textbf{Generate} step. 
Second, when $S^{'}$ comes from $V_s$, we need to fill the $[MASK]$ token with substitutes to construct a complete sentence. 
Follow \citet{li-etal-2020-bert-attack}, we use MLM \citep{devlin-etal-2019-bert} to generate k substitutes for the $[MASK]$         position in $S^{'}$ \footnote{We also experiment with other word substitution methods and find word substitutes generated from MLM work best.} and utilize WordNet \citep{wordnet} to filter out antonyms of original words. Then we iteratively substitute $[MASK]$ token with candidates in probability descending order computed by MLM and query the victim model for confidence scores. If one substitution successfully fools the victim model, we return the whole sentence as the final adversarial sample. Otherwise, we obtain the sentence, denoted as $S_w$, that causes the biggest drop in the victim model's confidence score. We compare $S_w$ with all candidates from $V_p$, choose the one that causes the biggest drop in the confidence score as the most potential candidate, and return to the \textbf{Generate} step.
\label{sec:maya}


\subsection{Combined with Behavior Cloning} 
As seen in Figure \ref{fig:pathdemo}, we use BERT$_{base}$ \citep{devlin-etal-2019-bert} and a linear classifier with one output unit as the architecture of MAYA$_\pi$. The core function of MAYA$_\pi$ is to predict the most potential candidate without querying the victim model.
In this section, We first describe how we exploit MAYA$_\pi$ to launch an adversarial attack because we require MAYA$_\pi$ to perform the full procedure of attacking in the training process. And then we detail the training process.

\subsubsection{Launch an Adversarial Attack}

Now assume that we have already trained an attack agent MAYA$_\pi$. Given the input sentence $S$, we follow the same procedure as in the \textbf{Generate} step in MAYA algorithm to generate adversarial candidate sets $V_p$ and $V_s$, corresponding to two different generation processes. Then, with the original sentence $S$ and an adversarial candidate $S_i$ concatenated as the input, MAYA$_\pi$ will output a score as a measure of tendency to choose this specific adversarial candidate. We obtain the candidate $S^{'}$ that get the highest score. Similarly, there are two different cases. 

First, when $S^{'}$ comes from $V_p$, we directly use $S^{'}$ to query the victim model. If it successfully fools the victim model, we return $S^{'}$ as the final adversarial sample. Otherwise, we view $S^{'}$ as the most potential candidate and return to the \textbf{Generate} step. 

Second, when $S^{'}$ comes from $V_s$, we follow the same procedure as in the \textbf{Pick} step in MAYA algorithm that iteratively substitutes $[MASK]$ token with candidate words and query the victim model for confidence scores. If one successful candidate found, we directly return this sentence as the final adversarial sample. Otherwise, we view the candidate that causes the biggest drop in the confidence score as the most potential candidate and return to the \textbf{Generate} step. The whole process will be repeated until a successful adversarial sample found or all potential candidates have been encountered before.
 In the next subsection, we will describe how we adapt this score-based \textbf{Pick} step to launch a decision-based black-box attack.

\label{sec:launch}  

\subsubsection{Training Process}

In this subsection, we describe our RL-based method to train MAYA$_\pi$ through Behavior Cloning with the expert knowledge from our MAYA algorithm. Specifically, we improve the training process by adapting the Dataset Aggregation (DAGGER) method \citep{ross2011reduction}. The training process incorporates three parts, namely initialization, sampling trajectories, and training.

\paragraph{Initialization} We initialize MAYA$_\pi$ with pre-trained weights from BERT \citep{devlin-etal-2019-bert} and a random initialized MLP. Also, we initialize an empty trajectory dataset $D$. 

\begin{table*}[]
\resizebox{\textwidth}{!}{
\begin{tabular}{ccccccccc}
\hline
Dataset & \multicolumn{1}{l}{\#Class} & \multicolumn{1}{l}{Avg Len} & Train & Dev & Test & \multicolumn{1}{l}{BiLSTM Acc} & \multicolumn{1}{l}{BERT Acc} & \multicolumn{1}{l}{RoBERTa Acc} \\ \hline
SST-2   & 2                          & 20.94                         & 6920                      & 872                     & 1821                     &            82.65	                    & 91.76                  & 94.89                  \\
MNLI    & 3                          & 23.28                       &  391075                  & 0                 & 9711                    &     70.44                         & 84.00           & 87.48                     \\
AG's News & 4                          & 38.7               & 96000                     & 24000                   & 7600                     &                 91.26               &          93.75                   &   94.14   \\ 
\hline                         
\end{tabular}
}

\caption{\label{dataset info}
Detailed information of datasets and corresponding victim models' accuracy. \#Class denotes the classification number. Avg Len means the average sentence length. Train, Dev, and Test represent the number of samples in the training, validation, and test datasets. BiLSTM Acc, BERT Acc, and RoBERTa Acc  denotes the original classification accuracy of each victim model.
}
\end{table*}

\paragraph{Sampling Trajectories}
To train MAYA$_\pi$, we need to interact with the victim model to obtain the training data. Specifically, we train a local victim model that has the same architecture with the target victim model, expecting to approximate the decision boundary of the target victim model\footnote{It is unpractical to query the target victim model thousands of times to train an agent that is employed to attack this victim model.}.

We sample a batch of original sentences. For each sentence $S_0$, we generate adversarial candidates $C_1, ..., C_k$. As in the \textbf{Verify} and \textbf{Pick} steps in MAYA algorithm, one specific candidate will be chosen as the final successful adversarial sample or the most potential candidate. We view the candidate chosen by MAYA algorithm as the ground truth label and add $((S_0, C_1, ..., C_k), label)$ to our trajectory dataset $D$.

To fully train an agent that can tackle different situations, we need a large dataset $D$. So, we adapt DAGGER method. Specifically, when receiving the ground truth label from MAYA algorithm, MAYA$_\pi$ doesn't take the golden action indicated by MAYA. It will take the action based on its own prediction. That is, MAYA$_\pi$ will predict which candidate will most confuse the victim model and follow its own procedure of launching an adversarial attack. The predicted candidate will be treated as $S_0$ and we will continue the same process of sampling trajectories to augment the dataset $D$.

\paragraph{Training}
Then we train MAYA$_\pi$ for only one epoch using the trajectory dataset $D$. We model the training task as a multi-class classification problem.
For each sample $((S_0, C_1, ..., C_k), label)$ drawn from $D$, we concatenate $S_0$ with each $C_i$. Then we input the $k$ concatenated sentences to MAYA$_\pi$ to get $k$ scores. We treat these $k$ scores as logits and use the cross-entropy loss to train MAYA$_\pi$. After the training process, we clear $D$ and continue the sampling procedure. 
The implementation details are described in Appendix \ref{sec:trainingdetails}.

\subsection{Adapted to Decision-based Attack}

To adapt MAYA$_\pi$ to decision-based attack, we only need to modify one step in the attack procedure described in Section~\ref{sec:launch} while keep other steps unchanged. Specifically, when the candidate $S^{'}$ that gets the highest score from MAYA$_\pi$ is from $V_s$, we iteratively substitute $[MASK]$ token with candidate words to generate adversarial candidates and query the victim model for decisions. If one candidate successfully flips the label, we treat it as the final adversarial sample. Otherwise, to generate adversarial samples more efficiently, we take the candidate whose sentence embedding has the lowest cosine similarity with the sentence embedding of the original sentence as the most potential candidate. Our intuition is that the candidate that least resembles the original sentences is more likely to be a successful adversarial sample.

\section{Experiments}
We conduct comprehensive experiments to evaluate our attack models on the tasks of sentiment analysis, natural language inference, and news classification.

\subsection{Datasets and Victim Models}
For sentiment analysis, we choose SST-2 \citep{socher-etal-2013-recursive}, a binary sentiment classification benchmark dataset. For natural language inference, we choose mismatched MNLI dataset \citep{williams-etal-2018-broad}. For news classification, we choose AG's News dataset. The models need to choose one of the four classes including World, Sports, Business, and Sci/Tech, given an instance in AG's News \citep{zhang2015character}.

We evaluate our attack models by attacking three victim models including BiLSTM \citep{schuster1997bidirectional}, BERT \citep{devlin-etal-2019-bert}, and RoBERTa \citep{liu2019RoBERTa}.
Details of the datasets and the classification accuracy of victim models are listed in Table~\ref{dataset info}.


\subsection{Attack Models}
We implement all baseline attack models using the NLP attack package TextAttack \citep{morris-etal-2020-textattack} and OpenAttack \citep{zeng2020openattack}.  

\subsection*{Score-based Attack Models}
We comprehensively compare our score-based attack models with five representative and strong score-based attack models including (1) GA+Embedding \citep{alzantot-etal-2018-generating}, (2) PWWS+Synonym \citep{ren-etal-2019-generating}, (3) PSO+Sememe \citep{zang-etal-2020-word}, (4) TextFooler \citep{Jin2019IsBR}, (5) BERT-Attack \citep{li-etal-2020-bert-attack}. Details of baseline models are listed in Appendix \ref{sec:baselinedetails} and we describe details of our attack models below.

 \paragraph{MAYA}
We select 3 open-source paraphrase models as the building blocks of MAYA. Specifically, we choose BaiDu translation API \footnote{\url{https://fanyi-api.baidu.com/}} to perform back translation of the original sentence, style-transfer based paraphrase model \citep{krishna-etal-2020-reformulating}, and T5 \citep{JMLR:v21:20-074} based paraphrase model \footnote{\url{https://huggingface.co/Vamsi/T5_Paraphrase_Paws}}.

\paragraph{MAYA$_{bt}$}
We observe from our preliminary experiments that only using back translation model can achieve comparable performance in most of the cases and be more computation efficient. So, we also implement MAYA using only back translation model, denoted as MAYA$_{bt}$.

\paragraph{MAYA$_{\pi}$}
Due to the similar performance of MAYA and MAYA$_{bt}$ most of the time, we train our attack agent through behavior cloning with the expert knowledge from MAYA$_{bt}$ in consideration of the efficiency of training and launching an adversarial attack.

\subsection*{Decision-based Attack Models}

We consider two decision-based baseline models including (1) GAHard \citep{maheshwary2020generating} and (2) SCPN \citep{iyyer-etal-2018-adversarial}. Details of baseline models are listed in Appendix \ref{sec:baselinedetails}. We conduct exhaustive experiments to compare our decision-based MAYA$_{\pi}^{*}$ with existing decision-based attack models.

\subsection{Experimental Settings}
\paragraph{Hyper-parameters}
For our attack models, we set the number of word substitutes k to 10. And for MAYA, to ensure the quality of successful adversarial samples, we discard adversarial samples with modification number larger than 8, 8, and 12 in SST-2, MNLI, and AG's News respectively due to the difference of average sentence length in three datasets. Besides, we also set a maximum query number restriction to 15,000 for all attack models in the decision-based black-box attack setting due to the computation and time budget.


\paragraph{Evaluation Metrics}
We evaluate the attack models considering their attack success rate, attack efficiency, and the quality of adversarial samples. (1) Attack success rate is defined as the percentage of adversarial samples that successfully fool the victim model. (2) Attack efficiency is defined as the average query number to the victim model of crafting an adversarial sample. (3) We use four different metrics, including grammaticality, fluency, validity, and naturality to evaluate adversarial samples' quality. Specifically, we use Language-Tool to calculate the relative increase rate of grammar errors, GPT-2 \citep{radford2019language} to compute adversarial samples' perplexity as a measure of fluency, and ask human annotators to evaluate adversarial samples' validity and naturality.


\useunder{\uline}{\ul}{}
\begin{table*}[]
\resizebox{\textwidth}{!}{
\begin{tabular}{cc|rrrr|rrrr|rrrr}
\hline
\rowcolor[HTML]{FFFFFF} 
\cellcolor[HTML]{FFFFFF}                                               & Victim Model                       & \multicolumn{4}{c|}{\cellcolor[HTML]{FFFFFF}BiLSTM}                                                                                                                                                              & \multicolumn{4}{c|}{\cellcolor[HTML]{FFFFFF}BERT}                                                                                                                                                                         & \multicolumn{4}{c}{\cellcolor[HTML]{FFFFFF}RoBERTa}                                                                                                                                                      \\ \cline{3-14} 
\rowcolor[HTML]{FFFFFF} 
\multirow{-2}{*}{\cellcolor[HTML]{FFFFFF}Dataset}                      & Attack Method                      & \multicolumn{1}{c}{\cellcolor[HTML]{FFFFFF}ASR}         & \multicolumn{1}{c}{\cellcolor[HTML]{FFFFFF}Query} & \multicolumn{1}{c}{\cellcolor[HTML]{FFFFFF}PPL} & \multicolumn{1}{c|}{\cellcolor[HTML]{FFFFFF}\%I} & \multicolumn{1}{c}{\cellcolor[HTML]{FFFFFF}ASR}            & \multicolumn{1}{c}{\cellcolor[HTML]{FFFFFF}Query}   & \multicolumn{1}{c}{\cellcolor[HTML]{FFFFFF}PPL}     & \multicolumn{1}{c|}{\cellcolor[HTML]{FFFFFF}\%I} & \multicolumn{1}{c}{\cellcolor[HTML]{FFFFFF}ASR}   & \multicolumn{1}{c}{\cellcolor[HTML]{FFFFFF}Query} & \multicolumn{1}{c}{\cellcolor[HTML]{FFFFFF}PPL} & \multicolumn{1}{c}{\cellcolor[HTML]{FFFFFF}\%I} \\ \hline
\rowcolor[HTML]{FFFFFF} 
\multicolumn{1}{c|}{\cellcolor[HTML]{FFFFFF}}                          & GA+Embedding                       & {\color[HTML]{000000} 75.91}                            & {\color[HTML]{000000} 833.18}                     & {\color[HTML]{000000} 523.89}                   & {\color[HTML]{000000} 4.45}                      & {\color[HTML]{000000} 53.33}                               & {\color[HTML]{000000} 1004.49}                      & {\color[HTML]{000000} 568.20}                       & {\color[HTML]{000000} 7.46}                      & {\color[HTML]{000000} 50.23}                      & {\color[HTML]{000000} 1040.03}                    & {\color[HTML]{000000} 664.43}                   & {\color[HTML]{000000} 7.91}                     \\
\rowcolor[HTML]{FFFFFF} 
\multicolumn{1}{c|}{\cellcolor[HTML]{FFFFFF}}                          & PWWS+Synonym                       & {\color[HTML]{000000} 87.75}                            & {\color[HTML]{000000} 130.68}                     & {\color[HTML]{000000} 658.68}                   & {\color[HTML]{000000} 9.05}                      & {\color[HTML]{000000} 75.12}                               & {\color[HTML]{000000} 145.45}                       & {\color[HTML]{000000} 837.91}                       & {\color[HTML]{000000} 8.35}                      & {\color[HTML]{000000} 77.03}                      & {\color[HTML]{000000} 149.31}                     & {\color[HTML]{000000} 1114.15}                  & {\color[HTML]{000000} 11.12}                    \\
\rowcolor[HTML]{FFFFFF} 
\multicolumn{1}{c|}{\cellcolor[HTML]{FFFFFF}}                          & PSO+Sememe                         & {\color[HTML]{000000} 90.44}                            & {\color[HTML]{000000} 2912.45}                    & {\color[HTML]{000000} 748.89}                   & {\color[HTML]{000000} 4.91}                      & {\color[HTML]{000000} 85.60}                               & {\color[HTML]{000000} 5354.05}                      & {\color[HTML]{000000} 759.30}                       & {\color[HTML]{000000} 4.96}                      & {\color[HTML]{000000} 85.50}                      & {\color[HTML]{000000} 5648.08}                    & {\color[HTML]{000000} 741.09}                   & {\color[HTML]{000000} 4.96}                     \\
\rowcolor[HTML]{FFFFFF} 
\multicolumn{1}{c|}{\cellcolor[HTML]{FFFFFF}}                          & TextFooler                         & {\color[HTML]{000000} 94.89}                            & {\color[HTML]{000000} 75.64}                      & {\color[HTML]{000000} 837.89}                   & {\color[HTML]{000000} 5.33}                      & {\color[HTML]{000000} 85.36}                               & {\color[HTML]{000000} 99.42}                        & {\color[HTML]{000000} 781.96}                       & {\color[HTML]{000000} 8.88}                      & {\color[HTML]{000000} 87.12}                      & {\color[HTML]{000000} 104.23}                     & {\color[HTML]{000000} 850.42}                   & {\color[HTML]{000000} 8.44}                     \\
\rowcolor[HTML]{FFFFFF} 
\multicolumn{1}{c|}{\cellcolor[HTML]{FFFFFF}}                          & BERT-Attack                        & {\color[HTML]{000000} \textbf{98.65}}                   & {\color[HTML]{000000} 52.76}                      & {\color[HTML]{000000} 675.75}                   & {\color[HTML]{000000} 10.15}                     & {\color[HTML]{000000} 90.36}                               & {\color[HTML]{000000} 81.44}                        & {\color[HTML]{000000} 686.76}                       & {\color[HTML]{000000} 15.32}                     & {\color[HTML]{000000} 95.01}                      & {\color[HTML]{000000} 81.67}                      & {\color[HTML]{000000} 683.64}                   & {\color[HTML]{000000} 13.82}                    \\
\rowcolor[HTML]{FFFFFF} 
\multicolumn{1}{c|}{\cellcolor[HTML]{FFFFFF}}                          & \cellcolor[HTML]{FFFFFF}MAYA$_{bt}$    & {\color[HTML]{000000} 97.71}                            & {\color[HTML]{000000} 61.46}                      & {\color[HTML]{000000} 454.18}                   & {\color[HTML]{000000} -6.11}                     & {\color[HTML]{000000} {\ul 97.83}}                         & {\color[HTML]{000000} 71.52}                        & {\color[HTML]{000000} \textbf{414.38}}              & {\color[HTML]{000000} -2.61}                     & {\color[HTML]{000000} 94.78}                      & {\color[HTML]{000000} 77.01}                      & {\color[HTML]{000000} 462.87}                   & {\color[HTML]{000000} -4.08}                    \\
\rowcolor[HTML]{FFFFFF} 
\multicolumn{1}{c|}{\cellcolor[HTML]{FFFFFF}}                          & MAYA                               & {\color[HTML]{000000} 98.52}                            & {\color[HTML]{000000} 59.11}                      & {\color[HTML]{000000} 411.67}                   & {\color[HTML]{000000} -6.11}                     & {\color[HTML]{000000} {\ul 98.81}}                         & {\color[HTML]{000000} 71.34}                        & {\color[HTML]{000000} 415.35}                       & {\color[HTML]{000000} \textbf{-4.16}}            & {\color[HTML]{000000} \textbf{96.64}}             & {\color[HTML]{000000} 75.38}                      & {\color[HTML]{000000} \textbf{437.32}}          & {\color[HTML]{000000} -6.18}                    \\
\rowcolor[HTML]{FFFFFF} 
\multicolumn{1}{c|}{\multirow{-8}{*}{\cellcolor[HTML]{FFFFFF}SST-2}}   & MAYA$_\pi$                         & {\color[HTML]{000000} 97.98}                            & {\color[HTML]{000000} \textbf{17.58}}             & {\color[HTML]{000000} \textbf{399.35}}          & {\color[HTML]{000000} \textbf{-7.62}}            & {\color[HTML]{000000} 94.40}                               & {\color[HTML]{000000} \textbf{18.32}}               & {\color[HTML]{000000} 435.86}                       & {\color[HTML]{000000} -4.14}                     & {\color[HTML]{000000} 95.24}                      & {\color[HTML]{000000} \textbf{19.71}}             & {\color[HTML]{000000} 471.92}                   & {\color[HTML]{000000} \textbf{-6.53}}           \\ \hline
\rowcolor[HTML]{FFFFFF} 
\multicolumn{1}{c|}{\cellcolor[HTML]{FFFFFF}}                          & GA+Embedding                       & {\color[HTML]{000000} 61.00}                            & {\color[HTML]{000000} 859.04}                     & {\color[HTML]{000000} 710.81}                   & {\color[HTML]{000000} 9.28}                      & {\color[HTML]{000000} 55.00}                               & {\color[HTML]{000000} 847.68}                       & {\color[HTML]{000000} 1080.79}                      & {\color[HTML]{000000} 10.98}                     & {\color[HTML]{000000} 41.80}                      & {\color[HTML]{000000} 857.52}                     & {\color[HTML]{000000} 847.06}                   & {\color[HTML]{000000} 10.63}                    \\
\rowcolor[HTML]{FFFFFF} 
\multicolumn{1}{c|}{\cellcolor[HTML]{FFFFFF}}                          & PWWS+Synonym                       & {\color[HTML]{000000} 77.30}                            & {\color[HTML]{000000} 156.53}                     & {\color[HTML]{000000} 2509.04}                  & {\color[HTML]{000000} 9.81}                      & {\color[HTML]{000000} 75.10}                               & {\color[HTML]{000000} 151.50}                       & {\color[HTML]{000000} 948.40}                       & {\color[HTML]{000000} 10.67}                     & {\color[HTML]{000000} 71.60}                      & {\color[HTML]{000000} 151.24}                     & {\color[HTML]{000000} 715.42}                   & {\color[HTML]{000000} 9.86}                     \\
\rowcolor[HTML]{FFFFFF} 
\multicolumn{1}{c|}{\cellcolor[HTML]{FFFFFF}}                          & PSO+Sememe                         & {\color[HTML]{000000} 80.80}                            & {\color[HTML]{000000} 5058.81}                    & {\color[HTML]{000000} 1208.33}                  & {\color[HTML]{000000} 5.20}                      & \cellcolor[HTML]{FFFFFF}{\color[HTML]{000000} 75.80}          & \cellcolor[HTML]{FFFFFF}{\color[HTML]{000000} 4491.1} & \cellcolor[HTML]{FFFFFF}{\color[HTML]{000000} 980.79} & \cellcolor[HTML]{FFFFFF}{\color[HTML]{000000} 4.91} & {\color[HTML]{000000} 76.10}                      & {\color[HTML]{000000} 5178.30}                    & {\color[HTML]{000000} 1075.73}                  & {\color[HTML]{000000} 5.01}                     \\
\rowcolor[HTML]{FFFFFF} 
\multicolumn{1}{c|}{\cellcolor[HTML]{FFFFFF}}                          & TextFooler                         & {\color[HTML]{000000} {\ul 87.10}}                      & {\color[HTML]{000000} 102.12}                     & {\color[HTML]{000000} 1715.32}                  & {\color[HTML]{000000} 10.27}                     & {\color[HTML]{000000} 84.70}                               & {\color[HTML]{000000} 96.26}                        & {\color[HTML]{000000} 2656.92}                      & {\color[HTML]{000000} 8.03}                      & {\color[HTML]{000000} 83.60}                      & {\color[HTML]{000000} 92.91}                      & {\color[HTML]{000000} 2703.12}                  & {\color[HTML]{000000} 8.51}                     \\
\rowcolor[HTML]{FFFFFF} 
\multicolumn{1}{c|}{\cellcolor[HTML]{FFFFFF}}                          & BERT-Attack                        & {\color[HTML]{000000} 84.90}                            & {\color[HTML]{000000} 81.15}                      & {\color[HTML]{000000} 5617.47}                  & {\color[HTML]{000000} 8.99}                      & {\color[HTML]{000000} 82.10}                               & {\color[HTML]{000000} 114.68}                       & {\color[HTML]{000000} 11761.92}                     & {\color[HTML]{000000} 16.07}                     & {\color[HTML]{000000} 84.10}                      & {\color[HTML]{000000} 66.13}                      & {\color[HTML]{000000} 10414.33}                 & {\color[HTML]{000000} 7.67}                     \\
\rowcolor[HTML]{FFFFFF} 
\multicolumn{1}{c|}{\cellcolor[HTML]{FFFFFF}}                          & \cellcolor[HTML]{FFFFFF}MAYA$_{bt}$    & {\color[HTML]{000000} 84.10}                            & {\color[HTML]{000000} 85.14}                      & {\color[HTML]{000000} 434.94}                   & {\color[HTML]{000000} -2.98}                     & {\color[HTML]{000000} 90.60}                               & {\color[HTML]{000000} 78.65}                        & {\color[HTML]{000000} 626.80}                       & {\color[HTML]{000000} -0.60}                     & {\color[HTML]{000000} 86.50}                      & {\color[HTML]{000000} 69.85}                      & {\color[HTML]{000000} \textbf{413.13}}          & {\color[HTML]{000000} -0.68}                    \\
\rowcolor[HTML]{FFFFFF} 
\multicolumn{1}{c|}{\cellcolor[HTML]{FFFFFF}}                          & MAYA                               & {\color[HTML]{000000} {\ul 87.10}}                      & {\color[HTML]{000000} 83.72}                      & {\color[HTML]{000000} \textbf{425.63}}          & {\color[HTML]{000000} -3.61}                     & {\color[HTML]{000000} \textbf{91.70}}                      & {\color[HTML]{000000} 76.00}                        & {\color[HTML]{000000} \textbf{599.44}}              & {\color[HTML]{000000} \textbf{-2.00}}            & {\color[HTML]{000000} \textbf{89.60}}             & {\color[HTML]{000000} 70.50}                      & {\color[HTML]{000000} 516.17}                   & {\color[HTML]{000000} \textbf{-3.89}}           \\
\rowcolor[HTML]{FFFFFF} 
\multicolumn{1}{c|}{\multirow{-8}{*}{\cellcolor[HTML]{FFFFFF}MNLI}}    & MAYA$_\pi$                         & \cellcolor[HTML]{FFFFFF}{\color[HTML]{000000} 81.20}       & {\color[HTML]{000000} \textbf{33.60}}             & {\color[HTML]{000000} 504.22}                   & {\color[HTML]{000000} \textbf{-3.69}}            & {\color[HTML]{000000} 85.40}                               & {\color[HTML]{000000} \textbf{26.55}}               & {\color[HTML]{000000} 719.42}                       & {\color[HTML]{000000} 2.50}                      & {\color[HTML]{000000} 84.60}                      & {\color[HTML]{000000} \textbf{26.93}}             & {\color[HTML]{000000} 654.00}                   & {\color[HTML]{000000} 0.32}                     \\ \hline
\rowcolor[HTML]{FFFFFF} 
\multicolumn{1}{c|}{\cellcolor[HTML]{FFFFFF}}                          & GA+Embedding                       & {\color[HTML]{000000} 67.40}                            & {\color[HTML]{000000} 1668.93}                    & {\color[HTML]{000000} 288.11}                   & {\color[HTML]{000000} 3.16}                      & {\color[HTML]{000000} 38.62}                               & {\color[HTML]{000000} 1490.97}                      & {\color[HTML]{000000} 399.80}                       & {\color[HTML]{000000} 7.77}                      & {\color[HTML]{000000} 34.33}                      & {\color[HTML]{000000} 1557.70}                    & {\color[HTML]{000000} 348.48}                   & {\color[HTML]{000000} 6.56}                     \\
\rowcolor[HTML]{FFFFFF} 
\multicolumn{1}{c|}{\cellcolor[HTML]{FFFFFF}}                          & PWWS+Synonym                       & {\color[HTML]{343434} 74.70}                            & {\color[HTML]{343434} 227.37}                     & {\color[HTML]{343434} 364.58}                   & {\color[HTML]{000000} 5.70}                      & {\color[HTML]{343434} 65.50}                               & {\color[HTML]{343434} 251.82}                       & {\color[HTML]{343434} 544.43}                       & {\color[HTML]{000000} 8.79}                      & {\color[HTML]{343434} 54.70}                      & {\color[HTML]{343434} 254.31}                     & {\color[HTML]{343434} 491.69}                   & {\color[HTML]{000000} 10.12}                    \\
\rowcolor[HTML]{FFFFFF} 
\multicolumn{1}{c|}{\cellcolor[HTML]{FFFFFF}}                          & \cellcolor[HTML]{FFFFFF}PSO+Sememe & {\color[HTML]{343434} 89.40}                            & {\color[HTML]{343434} 14028.64}                    & {\color[HTML]{343434} 548.63}                   & {\color[HTML]{343434} 10.01}                      & {\color[HTML]{343434} 66.20}                               & {\color[HTML]{343434} 15461.61}                      & {\color[HTML]{343434} 680.31}                       & {\color[HTML]{343434} 8.87}                      & {\color[HTML]{343434} 64.40}                      & {\color[HTML]{343434} 17048.21}                    & {\color[HTML]{343434} 639.70}                   & {\color[HTML]{343434} 9.15}                     \\
\rowcolor[HTML]{FFFFFF} 
\multicolumn{1}{c|}{\cellcolor[HTML]{FFFFFF}}                          & TextFooler                         & {\color[HTML]{000000} 84.20}                            & {\color[HTML]{000000} 163.36}                     & {\color[HTML]{000000} 335.13}                   & {\color[HTML]{000000} 3.32}                      & {\color[HTML]{000000} 88.70}                               & {\color[HTML]{000000} 215.61}                       & {\color[HTML]{000000} 600.53}                       & {\color[HTML]{000000} 8.79}                      & {\color[HTML]{000000} 78.20}                      & {\color[HTML]{000000} 232.08}                     & {\color[HTML]{000000} 562.13}                   & {\color[HTML]{000000} 6.47}                     \\
\rowcolor[HTML]{FFFFFF} 
\multicolumn{1}{c|}{\cellcolor[HTML]{FFFFFF}}                          & BERT-Attack                        & {\color[HTML]{000000} 88.40}                            & {\color[HTML]{000000} 187.18}                     & {\color[HTML]{000000} 369.47}                   & {\color[HTML]{000000} 4.20}                      & {\color[HTML]{000000} 81.30}                               & {\color[HTML]{000000} 180.51}                       & {\color[HTML]{000000} 375.41}                       & {\color[HTML]{000000} 9.52}                      & {\color[HTML]{000000} 82.60}                      & {\color[HTML]{000000} 206.59}                     & {\color[HTML]{000000} 394.50}                   & {\color[HTML]{000000} 3.51}                     \\
\rowcolor[HTML]{FFFFFF} 
\multicolumn{1}{c|}{\cellcolor[HTML]{FFFFFF}}                          & \cellcolor[HTML]{FFFFFF}MAYA$_{bt}$    & {\color[HTML]{000000} 84.60}                            & {\color[HTML]{000000} 215.61}                     & {\color[HTML]{000000} 206.47}                   & {\color[HTML]{000000} -5.64}                     & {\color[HTML]{000000} 82.60}                               & {\color[HTML]{000000} 248.61}                       & {\color[HTML]{000000} 237.63}                       & {\color[HTML]{000000} -11.36}                    & {\color[HTML]{000000} 78.10}                      & {\color[HTML]{000000} 234.29}                     & {\color[HTML]{000000} 221.82}                   & {\color[HTML]{000000} -6.15}                    \\
\rowcolor[HTML]{FFFFFF} 
\multicolumn{1}{c|}{\cellcolor[HTML]{FFFFFF}}                          & MAYA                               & {\color[HTML]{000000} {\ul 91.29}}                      & {\color[HTML]{000000} 183.46}                     & {\color[HTML]{000000} \textbf{172.25}}          & {\color[HTML]{000000} \textbf{-13.09}}           & \cellcolor[HTML]{FFFFFF}{\color[HTML]{000000} \textbf{93.10}} & {\color[HTML]{000000} 202.74}                       & {\color[HTML]{000000} \textbf{206.93}}              & {\color[HTML]{000000} -15.03}                    & {\color[HTML]{000000} \textbf{84.25}}             & {\color[HTML]{000000} 225.63}                     & {\color[HTML]{000000} \textbf{192.91}}          & {\color[HTML]{000000} -12.82}                   \\
\rowcolor[HTML]{FFFFFF} 
\multicolumn{1}{c|}{\multirow{-8}{*}{\cellcolor[HTML]{FFFFFF}AG's News}} & MAYA$_\pi$                         & \cellcolor[HTML]{FFFFFF}{\color[HTML]{343434} {\ul 90.40}} & {\color[HTML]{343434} \textbf{45.57}}             & {\color[HTML]{343434} 215.10}                   & {\color[HTML]{000000} -11.00}                    & \cellcolor[HTML]{FFFFFF}{\color[HTML]{343434} 81.00}          & {\color[HTML]{343434} \textbf{41.73}}               & {\color[HTML]{343434} 236.24}                       & {\color[HTML]{000000} \textbf{-17.45}}           & \cellcolor[HTML]{FFFFFF}{\color[HTML]{343434} 77.40} & {\color[HTML]{343434} \textbf{36.48}}             & {\color[HTML]{343434} 244.99}                   & {\color[HTML]{000000} \textbf{-17.08}}          \\ \hline
\end{tabular}

}

\caption{\label{Main experiments results}	
The results of attacking performance and adversarial samples' quality. ASR denotes the attack success rate. Query denotes the average query number of launching a successful adversarial attack. PPL and \%I indicate adversarial samples' fluency and relative increase of grammar errors. We also conduct Student's t-tests to measure the difference between different models. Boldfaced \textbf{numbers} mean significant advantage with p-value 0.05 as the threshold and underline {\ul numbers} mean no significant difference.
}
\end{table*}

\subsection{Experimental Results}
\paragraph{Attack Success Rate}
The attack success rate (ASR) results in score-based attack setting are listed in Table \ref{Main experiments results} and the results in decision-based setting are listed in Appendix \ref{sec:decision-based experiment results}.
Considering score-based attack, MAYA consistently outperforms all baseline models in three datasets and three victim models and MAYA$_\pi$ achieves comparable attack success rate with baseline models. In decision-based attack setting, MAYA$_{\pi}^{*}$ overall outperforms baseline models, especially in AG's News because the sentences in AG's News are much longer, providing more constituents to perturb. The results demonstrate the advantage of our multi-granularity attack models.

\begin{figure}[tbp]
\begin{center}
\includegraphics[width=0.5\textwidth]{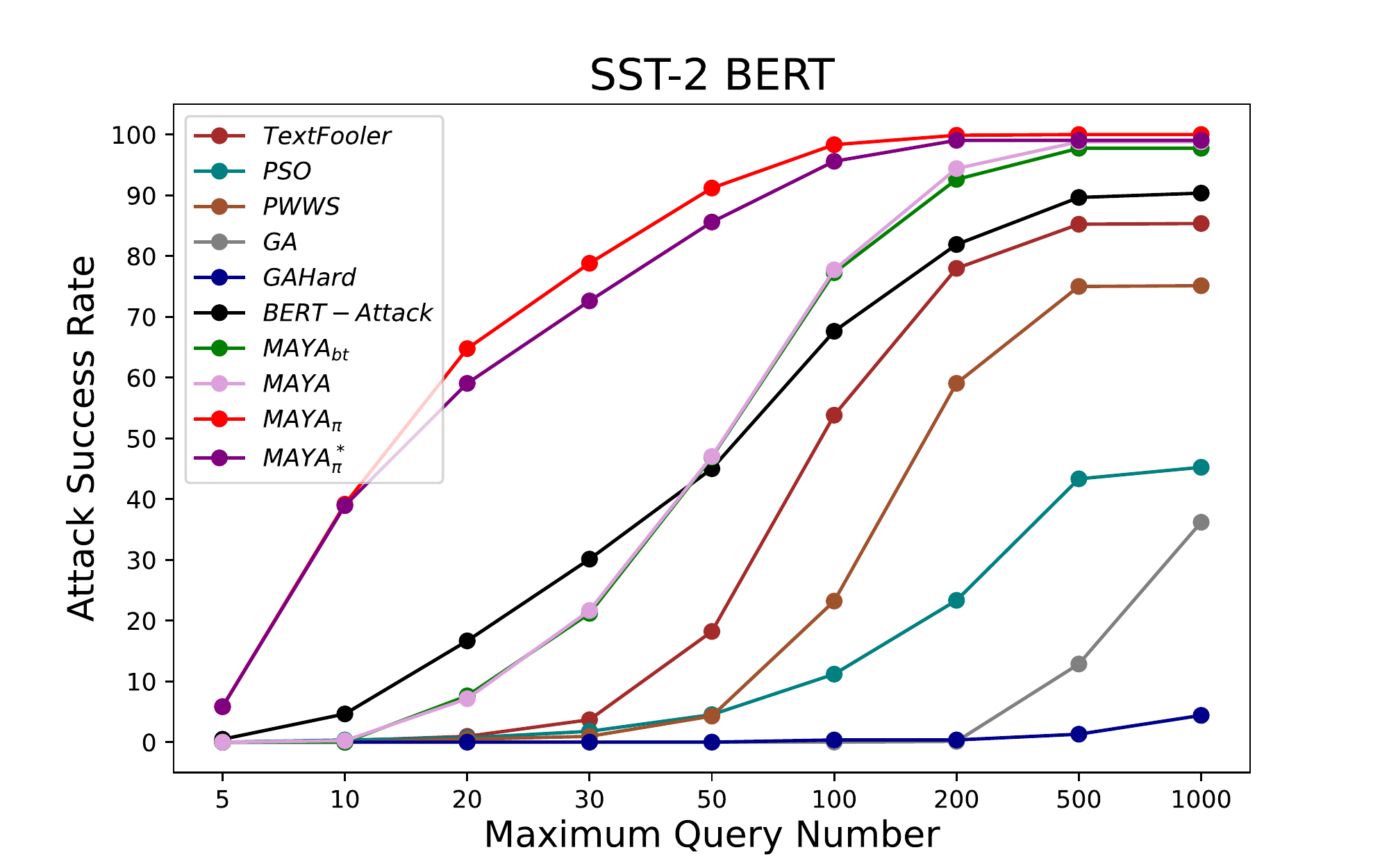}
\end{center}
\caption{Attack success rate under restriction of maximum query number in SST-2 when attacking BERT.}
\label{fig:query efficient}
\end{figure}

\begin{table}[]
\resizebox{0.95\linewidth}{!}{
\begin{tabular}{cc|rrrr}
\hline
                                             & Victim Model  & \multicolumn{4}{c}{BERT}                                                                                                                                                                                                   \\ \cline{3-6} 
\multirow{-2}{*}{SST-2}                      & Attack Method & \multicolumn{1}{c}{ASR}                              & \multicolumn{1}{c}{Query}                            & \multicolumn{1}{c}{PPL}                               & \multicolumn{1}{c}{I}                               \\ \hline
\multicolumn{1}{c|}{}                        & ADV           & \cellcolor[HTML]{FFFFFF}{\color[HTML]{000000} 18.81} & \cellcolor[HTML]{FFFFFF}{\color[HTML]{000000} 20.15} & \cellcolor[HTML]{FFFFFF}{\color[HTML]{000000} 411.78} & \cellcolor[HTML]{FFFFFF}{\color[HTML]{000000} -0.12}  \\
\multicolumn{1}{c|}{}                        & VERB          & \cellcolor[HTML]{FFFFFF}{\color[HTML]{000000} 27.02} & \cellcolor[HTML]{FFFFFF}{\color[HTML]{000000} 25.14} & \cellcolor[HTML]{FFFFFF}{\color[HTML]{000000} 346.05} & \cellcolor[HTML]{FFFFFF}{\color[HTML]{000000} 1.69}   \\
\multicolumn{1}{c|}{}                        & NOUN          & 37.14                                                & 40.80                                                 & 358.66                                                & 0.75                                                  \\
\multicolumn{1}{c|}{}                        & ADJ           & \cellcolor[HTML]{FFFFFF}{\color[HTML]{000000} 43.57} & \cellcolor[HTML]{FFFFFF}{\color[HTML]{000000} 22.89} & \cellcolor[HTML]{FFFFFF}{\color[HTML]{000000} 362.80}  & \cellcolor[HTML]{FFFFFF}{\color[HTML]{000000} 0.20}    \\
\multicolumn{1}{c|}{}                        & S             & \cellcolor[HTML]{FFFFFF}{\color[HTML]{000000} 12.86} & \cellcolor[HTML]{FFFFFF}{\color[HTML]{000000} \textbf{5.92}}  & \cellcolor[HTML]{FFFFFF}{\color[HTML]{000000} 316.15} & \cellcolor[HTML]{FFFFFF}{\color[HTML]{000000} \textbf{-22.10}} \\
\multicolumn{1}{c|}{}                        & VP            & 15.83                                                & 9.16                                                 & 357.01                                                & -9.20                                                 \\
\multicolumn{1}{c|}{}                        & NP            & \cellcolor[HTML]{FFFFFF}12.86                           & 12.61                                                & \textbf{298.38}                                                & -12.24                                                \\
\multicolumn{1}{c|}{}                        & BERT-Attack   & \cellcolor[HTML]{FFFFFF}{\color[HTML]{000000} 90.36} & 81.44                                                & 686.76                                                & 15.32                                                 \\
\multicolumn{1}{c|}{}                        & Phrases    & 51.90                                                & 25.91                                                & 381.63                                                & -13.74                                                \\
\multicolumn{1}{c|}{\multirow{-10}{*}{MAYA}} & All    & \cellcolor[HTML]{FFFFFF}\textbf{98.81}                      & \cellcolor[HTML]{FFFFFF}71.34                        & \cellcolor[HTML]{FFFFFF}415.35                        & \cellcolor[HTML]{FFFFFF}-4.16                         \\ \hline
\end{tabular}

}
\caption{\label{restrictionAttack}	
The results of attacking with restriction to the selection of different constituent types in SST-2.
}

\end{table}

\paragraph{Attack Efficiency}
For score-based attack, our models especially MAYA$_\pi$ show great superiority over all baseline models. For decision-based attack, MAYA$_{\pi}^{*}$ significantly outperforms GAHard, which needs thousands of queries.

Furthermore, we measure the attack success rate of attack models under the restriction of maximum query number. Figure \ref{fig:query efficient} show all attack models' attack success rate in SST-2 when attacking BERT under the restriction of maximum query number. 
Appendix \ref{sec:efffigs} shows remainder results in three datasets and three victim models. 
We observe that our RL-based attack models significantly outperform all baseline models under the restriction of maximum query number, demonstrating the practicality of our attack models in real-world situations. To confirm our argument, we present results of attacking two open-source NLP frameworks in Appendix \ref{sec:opensourceNLP}.

\paragraph{Adversarial Sample Quality}
We can observe from the results that our multi-granularity attack models overall outperform all baseline models considering adversarial samples' fluency and relative increase of grammar errors. Human evaluation results presented in Appendix \ref{sec:humaneval} also prove the high quality of our adversarial samples. 


\label{sec:experiments}

\section{Further Analysis}

\subsection{Constituent Selection}
It's important to investigate which constituent type our multi-granularity attack models tend to select as the vulnerable part of the sentence and the impact of different constituent types. We first investigate the selection frequency of all constituent types. The results are listed in Appendix \ref{sec:selectfreq}. Then, we select 7 constituent types that are more common and often selected as the vulnerable parts of sentences. We restrict the selection of MAYA to each of these constituent types and evaluate the attacking performance. We also list the results of attacking with restriction to only words (BERT-Attack), with restriction to only phrases (Phrases), and with no restriction (All) for comparison\footnote{We refer readers to \citep{taylor2003penn} for the meaning of syntax tags.}.

We can conclude from Table \ref{restrictionAttack} that while word-level substitution (BERT-Attack) ensures the attacking performance, there still exists a significant gap between word-level attack and our multi-granularity attack (All). Besides, paraphrasing constituent types improves the quality of sentences due to our strict restrictions and can produce adversarial samples with some probability.

\begin{table}[]
\resizebox{0.95\linewidth}{!}{
\begin{tabular}{cc|cc}
\hline
                                             & Victim Model  &                                                                            &                                                        \\
\multirow{-2}{*}{SST-2}                      & Attack Method & \multirow{-2}{*}{BiLSTM}                                                   & \multirow{-2}{*}{RoBERTa}                              \\ \hline
\multicolumn{1}{c|}{}                        & GA+Embedding  & \cellcolor[HTML]{FFFFFF}{\color[HTML]{000000} 28.37}                     & \cellcolor[HTML]{FFFFFF}{\color[HTML]{000000} 23.73} \\
\multicolumn{1}{c|}{}                        & PWWS+Synonym  & \cellcolor[HTML]{FFFFFF}{\color[HTML]{000000} 28.07}                     & \cellcolor[HTML]{FFFFFF}{\color[HTML]{000000} 21.66} \\
\multicolumn{1}{c|}{}                        & PSO+Sememe    & \cellcolor[HTML]{FFFFFF}{\color[HTML]{000000} 38.83}                     & \cellcolor[HTML]{FFFFFF}{\color[HTML]{000000} 40.22} \\
\multicolumn{1}{c|}{}                        & TextFooler    & \cellcolor[HTML]{FFFFFF}{\color[HTML]{000000} 28.38} & \cellcolor[HTML]{FFFFFF}{\color[HTML]{000000} 20.77} \\
\multicolumn{1}{c|}{}                        & BERT-Attack   & 28.12                                                                    & 29.19                                                \\
\multicolumn{1}{c|}{}                        & MAYA$_{bt}$        & 35.46                                                                    & 44.39                                                \\
\multicolumn{1}{c|}{}                        & MAYA          & \cellcolor[HTML]{FFFFFF}34.81                                            & \cellcolor[HTML]{FFFFFF}43.53                        \\
\multicolumn{1}{c|}{}                        & MAYA$_\pi$        & \textbf{40.70}                                                                    & \textbf{52.86}                                                \\
\multicolumn{1}{c|}{}                        & GAHard       & 25.47                                                                    & 16.55                                                \\
\multicolumn{1}{c|}{\multirow{-10}{*}{BERT}} & MAYA$_{\pi}^{*}$       & 39.17                                                                    & 39.17                                                \\ \hline
\end{tabular}

}

\caption{\label{transferability}	
The transfer attack success rate of adversarial samples in SST-2.
}

\end{table}

\subsection{Transferability}
We investigate the transferability of adversarial samples produced by all attack models in SST-2 with BERT as the victim model. We don't consider SCPN in this transferability study because this method is model agnostic and cannot be directly compared with other attack models. We can observe from Table \ref{transferability} that our MAYA$_\pi$ attack agent crafts adversarial samples with significantly higher transferability. That's probably because MAYA$\pi$ perturbs the sentence based not only on the outputs of victim models but also on its own prediction, ensuring adversarial samples to capture common vulnerabilities of different victim models.
\begin{table}[]
\resizebox{0.95\linewidth}{!}{
\begin{tabular}{cc|cccc}
\hline
\multirow{2}{*}{Victim Model} & Dataset       & \multicolumn{4}{c}{SST-2}                                          \\
                              & Attack Method & ASR            & Query          & PPL             & \%I            \\ \hline
\multirow{2}{*}{BERT}         & MAYA$_{r}$        & 74.17          & 30.90          & \textbf{398.51} & \textbf{-8.08} \\
                              & MAYA$_{\pi}$          & \textbf{94.40} & \textbf{18.32} & 435.86          & -4.14          \\ \hline
\end{tabular}

}
\caption{\label{ImpactImitation}	
The attacking performance of random initialized MAYA$_{\pi}$ and our MAYA$_{\pi}$ trained through behavior cloning.
}

\end{table}

\subsection{Impact of Imitation Algorithm}
Despite the strong attacking performance of MAYA$_{\pi}$, the impact of our adapted imitation algorithm is unknown. One may attribute the success to the capacity of the multi-granularity attack model, ignoring the contribution of the imitation learning process. So, we investigate the impact of our adapted imitation algorithm in this section. We employ a random initialized MAYA$_{\pi}$ without interacting with local victim models to launch attacks against BERT in SST-2. From Table~\ref{ImpactImitation}, we can conclude that the imitation learning process do bring some useful knowledge to our attack agent.

\begin{table}[]
\resizebox{0.48\textwidth}{!}{
\begin{tabular}{c
>{\columncolor[HTML]{FFFFFF}}c |
>{\columncolor[HTML]{FFFFFF}}c 
>{\columncolor[HTML]{FFFFFF}}c 
>{\columncolor[HTML]{FFFFFF}}c 
>{\columncolor[HTML]{FFFFFF}}c }
\hline
\cellcolor[HTML]{FFFFFF}                                                 & \cellcolor[HTML]{FFFFFF}Dataset       & \multicolumn{4}{c}{\cellcolor[HTML]{FFFFFF}SST-2}                                                                  \\ \cline{3-6} 
\multirow{-2}{*}{\cellcolor[HTML]{FFFFFF}Victim Model}                   & \cellcolor[HTML]{FFFFFF}Attack Method & \cellcolor[HTML]{FFFFFF}ASR & Query                      & PPL                         & \%I                        \\ \hline
\multicolumn{1}{c|}{\cellcolor[HTML]{FFFFFF}}                         & MAYA$_{\pi}^{o}$           & 97.98                     & 17.58                      & 399.35                      & -7.62                      \\
\multicolumn{1}{c|}{\multirow{-2}{*}{\cellcolor[HTML]{FFFFFF}BiLSTM}} & MAYA$_{\pi}^{B}$           & 92.87                     & \cellcolor[HTML]{FFFFFF}20.14 & \cellcolor[HTML]{FFFFFF}434.99 & \cellcolor[HTML]{FFFFFF}-5.18 \\ \hline
\multicolumn{1}{c|}{}                                                 & MAYA$_{\pi}^{o}$             & 95.24                   & \cellcolor[HTML]{FFFFFF}19.71 & \cellcolor[HTML]{FFFFFF}471.92 & \cellcolor[HTML]{FFFFFF}-6.53 \\
\multicolumn{1}{c|}{\multirow{-2}{*}{RoBERTa}}                        & MAYA$_{\pi}^{B}$                & 94.90                     & \cellcolor[HTML]{FFFFFF}18.51 & \cellcolor[HTML]{FFFFFF}425.16 & \cellcolor[HTML]{FFFFFF}-5.17 \\ \hline
\end{tabular}
}
\caption{\label{limitation}	
The results of attacking performance when the victim model's architecture is unkown. 
}

\end{table}

\subsection{Limitation}
We make a strong assumption in the development of our imitation algorithm that we assume that we have already known the victim models' architectures, which is unrealistic in real-world situations. However, as mentioned in Section~\ref{sec:experiments}, our attack agent can successfully launch adversarial attacks against real-world NLP frameworks, which confirms the practicability of our imitation algorithm. The results are presented in Appendix \ref{sec:opensourceNLP}. Further, we investigate the attacking performance of our attack agents in SST-2 dataset when the victim model's architecture is unknown. Specifically, we employ the attack agent trained by interacting with our local BERT model, denoted as MAYA$_{\pi}^{B}$  to launch adversarial attacks against BiLSTM and RoBERTa. We compare the attacking performance with that of originally trained attack agent, denoted as MAYA$_{\pi}^{o}$.

We can observe from Table~\ref{limitation} that MAYA$_{\pi}^{B}$ achieves similar attacking performance while maintain the attack efficiency and quality of adversarial samples, especially in RoBERTa. That's probably because the common features and architectures shared by pre-trained models, which strongly supports our view that our attack agents can indeed cause significant drop in models' prediction accuracy even though we the victim models' architectures are unknown because we can just assume the black-box system is built on pre-trained models, which can be verified in most cases.

\section{Conclusion and Future Work}
In this paper, we propose a multi-granularity adversarial attack model (MAYA) and propose a RL-based method to train an attack agent (MAYA$_\pi$) through behavior cloning with the expert knowledge from our MAYA algorithm. Further, we adapt MAYA$_{\pi}$ to decision-based attack setting, handling the issues of attacking models that only output decisions.
Experimental results show that our attack models achieve overall higher attacking performance and produce more fluent and grammatical adversarial samples. We also show that MAYA$_{\pi}$ can launch adversarial attacks towards open-source NLP frameworks, demonstrating the practicability of our attack agent in real-world situations.

In the future, we will focus on how to improve current models' robustness towards multi-granularity attacks. In addition, we will try to apply MAYA$_{\pi}$ to other less investigated settings in textual adversarial attack, such as to launch target attacks in decision-based attack setting.

\section{Acknowledgements}
This work was supported in part by the National Natural Science Foundation of China under Grant 61602197. Also, we thank all the anonymous reviewers for their valuable comments and suggestions.

\section{Ethical Considerations}
In this section, we discuss the potential broader impact and ethical considerations of our paper. 

\textbf{Intended use.}
In this paper, we propose multi-granularity attacking models that can handle different attack settings with superior performance. Our motivations are twofold. First, we can find some insights from the experimental results about current black-box machine learning models that can help us move towards explainable AI. Second, we demonstrate the potential risks of deploying current models in the real world, encouraging the research community to develop more robust models.

\textbf{Potential risk.}
It's possible that our attacking models may be maliciously used to launch an adversarial attack against off-the-shelf commercial systems. However, according to the research on adversarial attack on computer vision, it's important to make the research community realize these powerful attacking models before defending them. So, studying and investigating adversarial attack is significant.

\textbf{Energy saving.}
We present the details of our training process in Appendix J to prevent people from making unnecessary hyper-parameters tuning and help researchers to quickly reproduce our results. We will also release the checkpoints including all victim models and our attack agents to avoid energy costs to re-train them.


\bibliography{MAYAEMNLP.bib}
\bibliographystyle{acl_natbib}

\appendix
\section{MAYA Algorithm}
The whole process of MAYA Algorithm is shown in Algorithm \ref{mayaalgorithm}.

\label{sec:appendix}
\begin{algorithm*}[!h]
\caption{~MAYA}
\textbf{Input:} Original sentence $S$; Victim Model $M$\\
\textbf{Output:} Adversarial sample $S_{a}$
	
\begin{algorithmic}[1]

\STATE \textbf{Generate Procedure}  \\

\STATE \quad // generate candidates by paraphrasing constituent parts
\STATE \quad get constituent parts list $C$ $=$ ConstituentParser($S$)
\STATE \quad \textbf{for} each constituent part $c$ in $C$ \textbf{do}
\STATE \quad \quad get paraphrases list $P_c =ParaphraseModels(c)$ ~~ // use several paraphrase models
\STATE \quad  get adversary candidates set $V_p = \cup{filter(P_c)}$ ~~ // based on semantics and grammar rules

\STATE \quad // generate candidates by masking original words\\
\STATE \quad \textbf{for} each word $w$ in $S$:
\STATE \quad \quad mask $w$ to get $S_{w_i} = [w_0, ..., [MASK], ..., w_n]$ 
\STATE \quad get adversary candidates set $V_s = \{S_{w_1}, ..., S_{w_i}, ..., S_{w_n}\}$

\STATE \textbf{Verify Procedure} \\
\STATE\quad query the victim model for $scores = M(V_p \cup V_s)$

\STATE\quad \textbf{if} exist candidates list $L \subseteq (V_p \cup V_s)$ successfully fool M \textbf{then}
\STATE \quad \quad // we detail this process in 
\STATE \quad \quad return the most appropriate adversary sentence $S_a \in L$ or go to the \textbf{Pick Procedure}

\STATE \quad \textbf{else}
\STATE \quad \quad go to the \textbf{Pick Procedure}

\STATE \textbf{Pick Procedure} 
\STATE \quad select the most confusing candidate $S_a = Select(V_p \cup V_s, scores)$
\STATE \quad \textbf{if} $S_a \in V_p$ \textbf{then}
\STATE \quad \quad return $S_a$
\STATE \quad \textbf{else}
\STATE \quad \quad get substitute words list $W=MLM(S_a)$ // $S_a \in V_s$ and we choose k substitute words 

\STATE \quad \quad get complete sentences list $L_s=Fill(S_a, W)$
\STATE \quad \quad fill the $[MASK]$ token and query the victim model for $w\_scores = M(L_s)$
\STATE \quad \quad \textbf{if} exist $s \in L_s$  successfully fool the model \textbf{then}
\STATE \quad \quad \quad return $s$
\STATE \quad \quad \textbf{else}
\STATE \quad \quad \quad obtain the most confusing sentence $s_p$ from $V_p$

\STATE \quad \quad compare and return the most potential candidate $S_a=Choose(L_s, s_p, scores, w\_scores)$

\end{algorithmic}
\label{mayaalgorithm}
\end{algorithm*}

\begin{table*}[]
\resizebox{\textwidth}{!}{
\begin{tabular}{|l|l|l|l|l|l|l|l|l|l|l|l|l|l|l|}
\hline
\multicolumn{15}{|c|}{Sample I}                                                                                                                                                                                                                                                                                                                        \\ \hline
\multicolumn{15}{|l|}{Original Sentence}                                                                                                                                                                                                                                                                                                               \\ \hline
\multicolumn{15}{|l|}{\multirow{2}{*}{\begin{tabular}[c]{@{}l@{}}The second-ranked Jayhawks can redeem themselves for one of their most frustrating losses last season Monday  when \textcolor{blue}{they welcome the} \\ \textcolor{blue}{Wolf Pack to Allen Fieldhouse} .  \end{tabular}}} \\
\multicolumn{15}{|l|}{}                                                                                                                                                                                                                                                                                                                                \\ \hline
\multicolumn{15}{|l|}{Adversarial Sentence}                                                                                                                                                                                                                                                                                                            \\ \hline
\multicolumn{15}{|l|}{\begin{tabular}[c]{@{}l@{}}The second-ranked Jayhawks can redeem themselves for one of their most frustrating losses last season Monday when \textcolor{red}{the Wolf Pack is} \\ \textcolor{red}{welcomed  to the  semifinals} . \end{tabular}}                           \\ \hline
\multicolumn{15}{|c|}{Sample II}                                                                                                                                                                                                                                                                                                                       \\ \hline
\multicolumn{15}{|l|}{Original Sentence}                                                                                                                                                                                                                                                                                                               \\ \hline
\multicolumn{15}{|l|}{\multirow{2}{*}{\begin{tabular}[c]{@{}l@{}}If the playoffs opened right now, instead of next month, the A \#39 \textcolor{blue}{;} s would face the Red Sox in the first round -- \textcolor{blue}{again}.  Boston bounced  \\ Oakland  out of the postseason in five games last year, \textcolor{blue}{coming back from a 2-0 deficit to do so} .\end{tabular}}}                          \\
\multicolumn{15}{|l|}{}                                                                                                                                                                                                                                                                                                                                \\ \hline
\multicolumn{15}{|l|}{Adversarial Sentence}                                                                                                                                                                                                                                                                                                            \\ \hline
\multicolumn{15}{|l|}{\begin{tabular}[c]{@{}l@{}}If the playoffs opened right now, instead of next month, the A \#39 \textcolor{red}{.} s would face the Red Sox in the first round -- \textcolor{red}{once again} .  Boston \\ bounced Oakland out of the postseason in five games last year , \textcolor{red}{coming back from a deficit} .\end{tabular}}                                                \\ \hline
\end{tabular}

}

\caption{\label{casestudytable}	
Adversarial samples crafted by our multi-granularity attack models in AG’s News.
}

\end{table*}

\label{sec:mayaalgorithm}
\section{Details of Baseline Attack Models}
We describe details of baseline models in this section.

\paragraph{TextFooler} This model \citep{Jin2019IsBR} is a score-based attack method ranking words in the input sentence by saliency and chooses substitutes based on the word embedding to construct an adversarial sample.

\paragraph{PWWS+Synonym} This model \citep{ren-etal-2019-generating} is a greedy algorithm using augmented word saliency to iteratively substitute words with synonyms from WordNet \citep{wordnet}. 

\paragraph{GA+Embedding} This model \citep{alzantot-etal-2018-generating} uses classical genetic algorithm to search for an adversarial sample. It relies on word embeddings similarity to select word substitutions.

\paragraph{PSO+Sememes} This model \citep{zang-etal-2020-word} considers adversarial attack as a combinatorial optimization problem, and then designs an algorithm based on particle swarm optimization to substitute a word with its synonyms from HowNet \citep{dong-etal-2010-hownet}.

\paragraph{BERT-Attack} This model \citep{li-etal-2020-bert-attack} iteratively masks each word in original sentences, obtain substitutes from MLM and greedily select the substitute that causes the biggest drop in victim models' confidence scores.

\paragraph{SCPN} This model \citep{iyyer-etal-2018-adversarial} predefines several syntax patterns and paraphrase the original sentence to different syntax structure, intending to find an adversarial sample.

\paragraph{GAHard} This model \citep{maheshwary2020generating} involves initializing an potential adversarial sample, search space reduction, and population based optimization to find a semantic preserved adversarial sample.

\label{sec:baselinedetails}

%
%
%
%

\begin{table*}[]
\resizebox{\textwidth}{!}{
\begin{tabular}{c|ccccccccccccccc}
\hline
Constituent Type & ADJ  & NOUN  & VERB & VP    & S    & NP   & ADV & PP  & PUNCT & PRON & ADJP & ADP & CCONJ & PART & DET \\ \hline
Total Num   & 287  & 223   & 175  & 137   & 129  & 113  & 92  & 56  & 54    & 35   & 34   & 33  & 28    & 28   & 24  \\ \hline
Constituent Type & SBAR & SCONJ & AUX  & PROPN & FRAG & ADVP & NUM & NAC & INTJ  & WHPP & NX   & SQ  & PRN   & X    &     \\ \hline
Total Num   & 19   & 19    & 15   & 9     & 9    & 6    & 2   & 2   & 1     & 1    & 1    & 1   & 1     & 1    &     \\ \hline
\end{tabular}

}
\caption{\label{constituentselection}	
The frequency of selection of all constituents in SST-2.
}

\end{table*}

\section{Decision-based Experiment Results}
The decision-based experiment results are shown in Table~\ref{Main experiments results2}. Note that the number of syntax templates is fixed in SCPN, making the query number constant. So, the query number of SCPN cannot be directly compared with that of other decision-based attack models and we leave out all query number results of SCPN in Table~\ref{Main experiments results2}.

\label{sec:decision-based experiment results}

\section{Attack Efficiency}
\begin{figure*}[ht]
 \centering
 \begin{minipage}[b]{0.48\linewidth}
   \includegraphics[height=5.5cm,width=8cm]{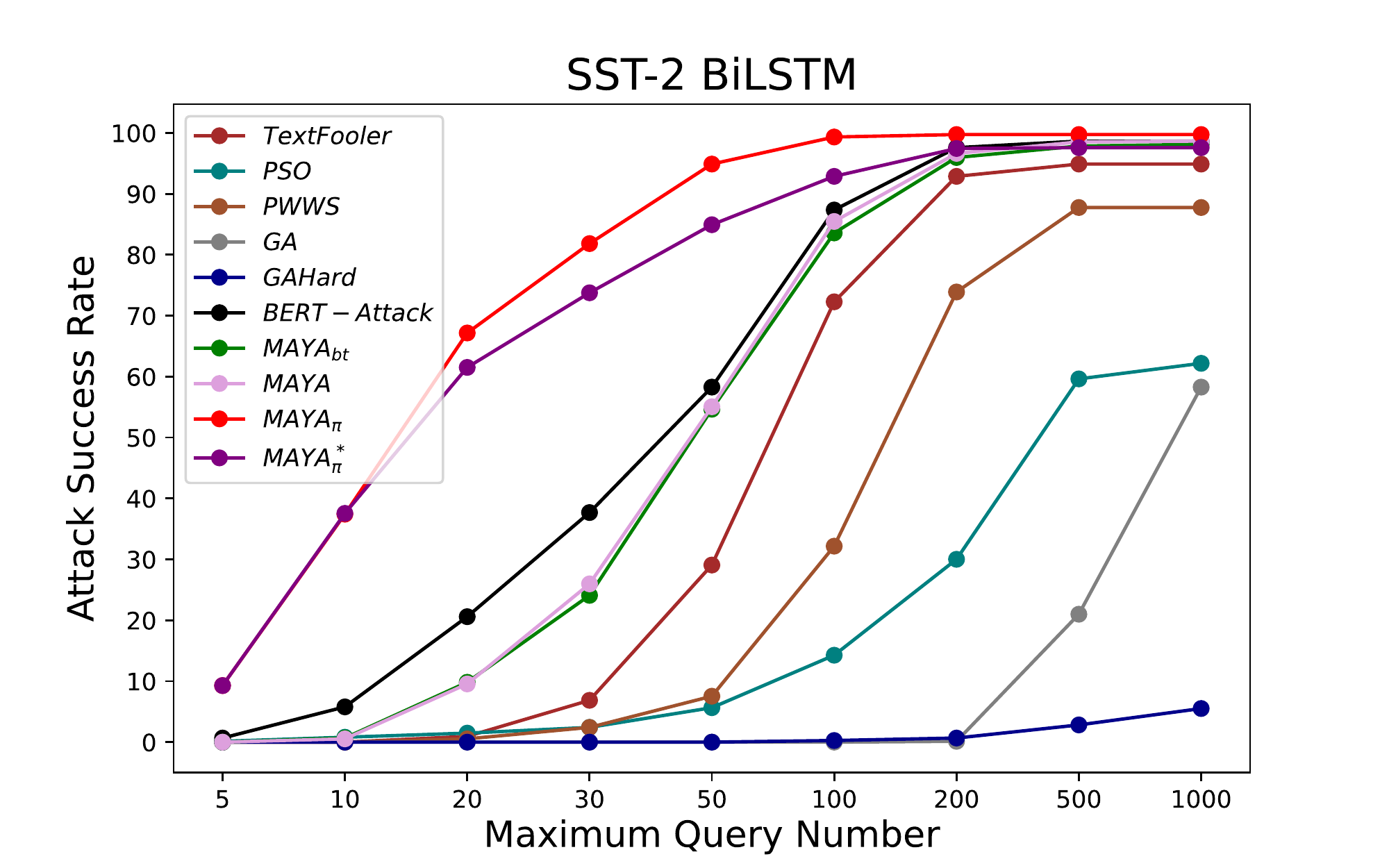}
   \caption{Attack BiLSTM in SST-2}
   \label{fig:minipage1}
 \end{minipage}
 \quad
 \begin{minipage}[b]{0.48\linewidth}
    \includegraphics[height=5.5cm,width=8cm]{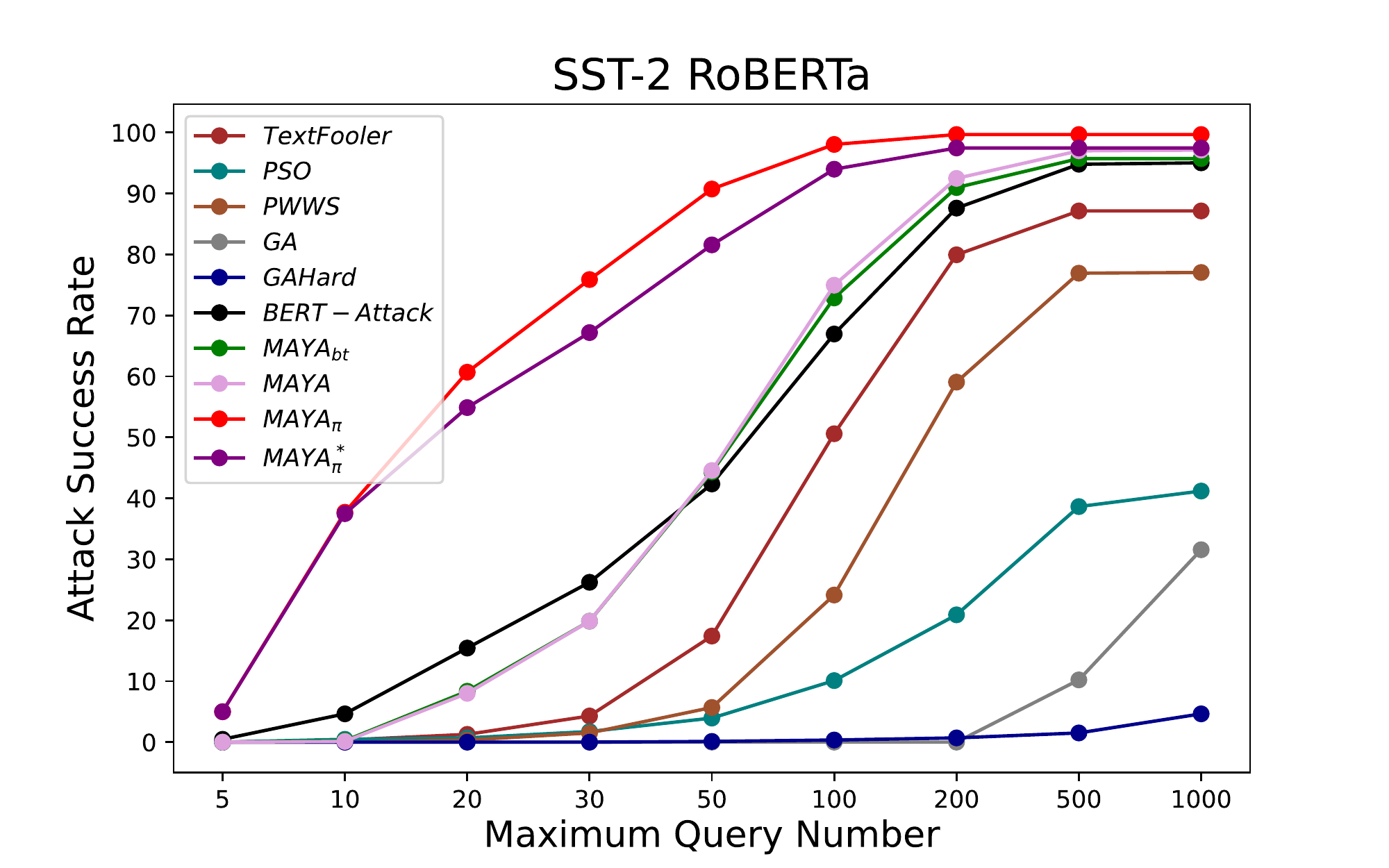}
      \caption{Attack RoBERTa in SST-2}
  \label{fig:minipage2}
\end{minipage}

 \begin{minipage}[b]{0.48\linewidth}
   \includegraphics[height=5.5cm,width=8cm]{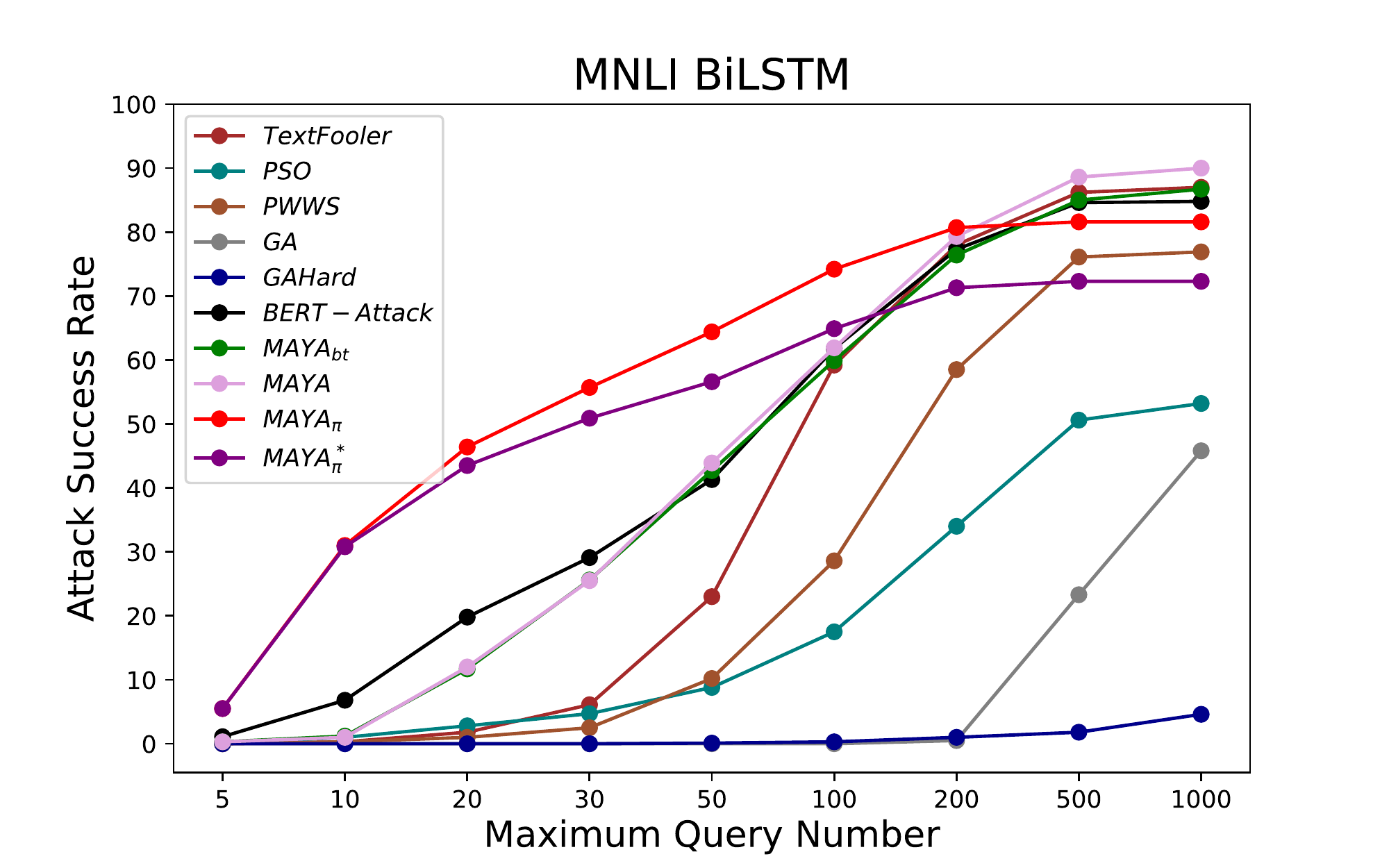}
   \caption{Attack BiLSTM in MNLI}
   \label{fig:minipage3}
 \end{minipage}
 \quad
 \begin{minipage}[b]{0.48\linewidth}
    \includegraphics[height=5.5cm,width=8cm]{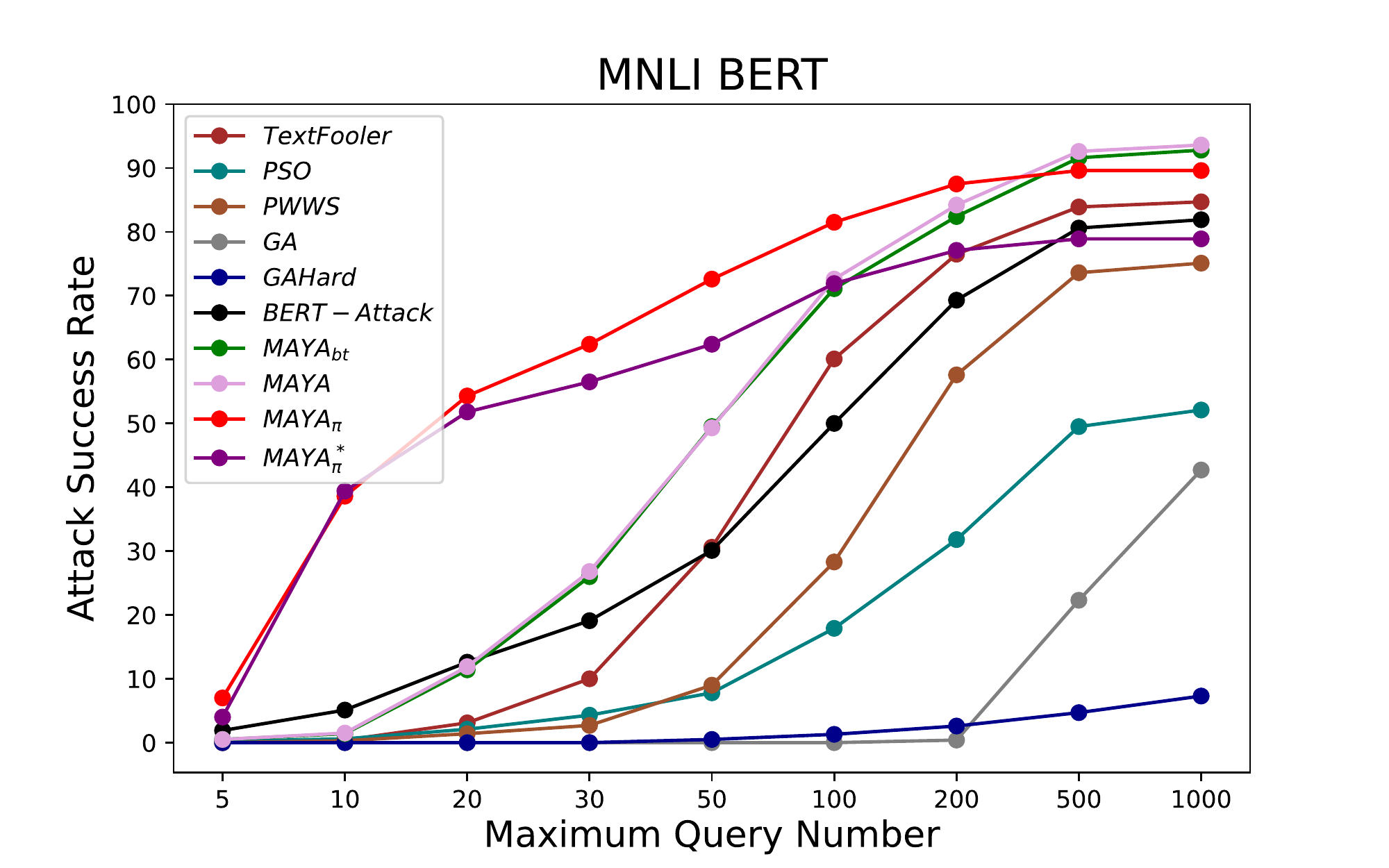}
      \caption{Attack BERT in MNLI}
  \label{fig:minipage4}
\end{minipage}

 \begin{minipage}[b]{0.48\linewidth}
   \includegraphics[height=5.5cm,width=8cm]{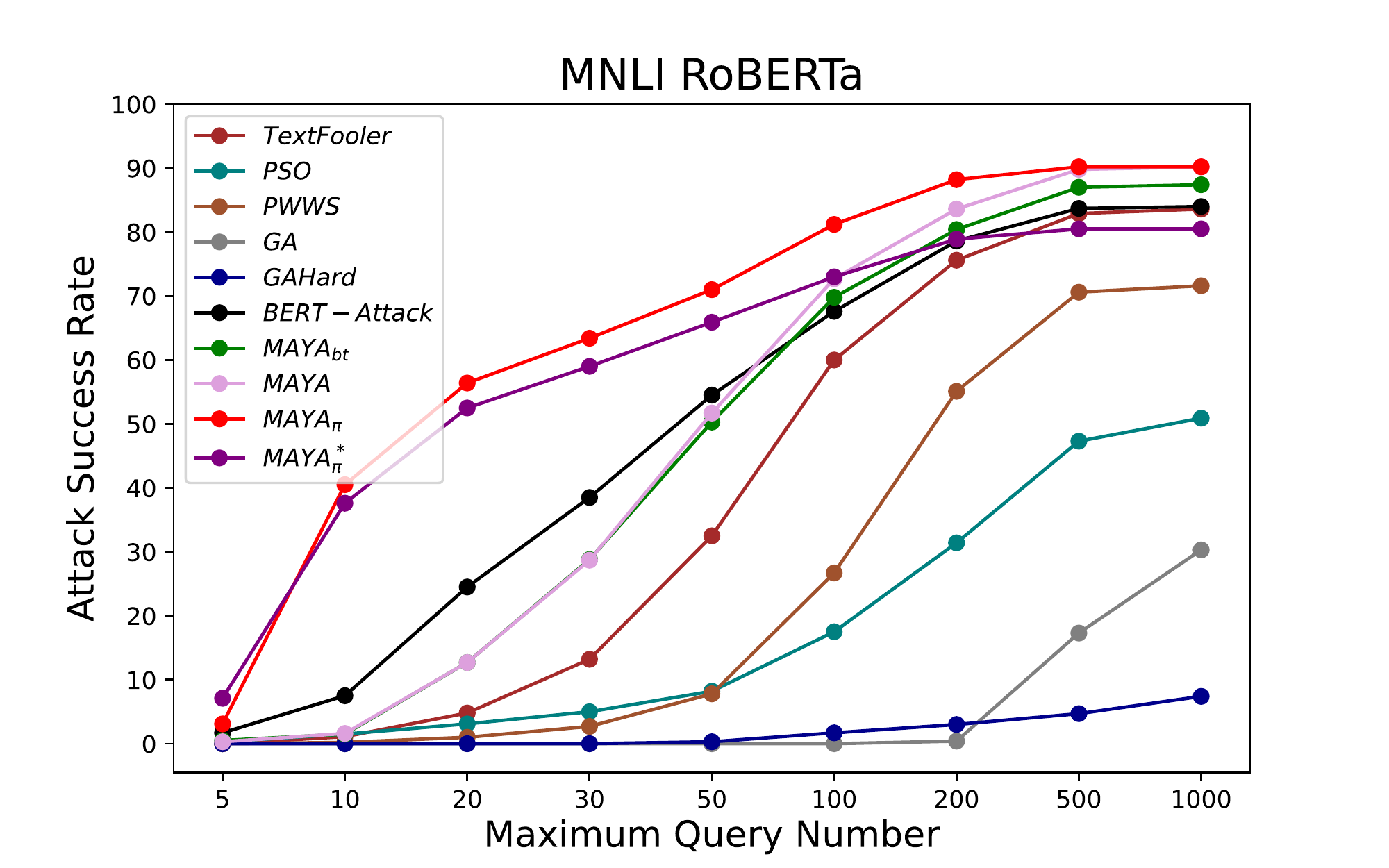}
   \caption{Attack RoBERTa in MNLI}
   \label{fig:minipage5}
 \end{minipage}
 \quad
 \begin{minipage}[b]{0.48\linewidth}
    \includegraphics[height=5.5cm,width=8cm]{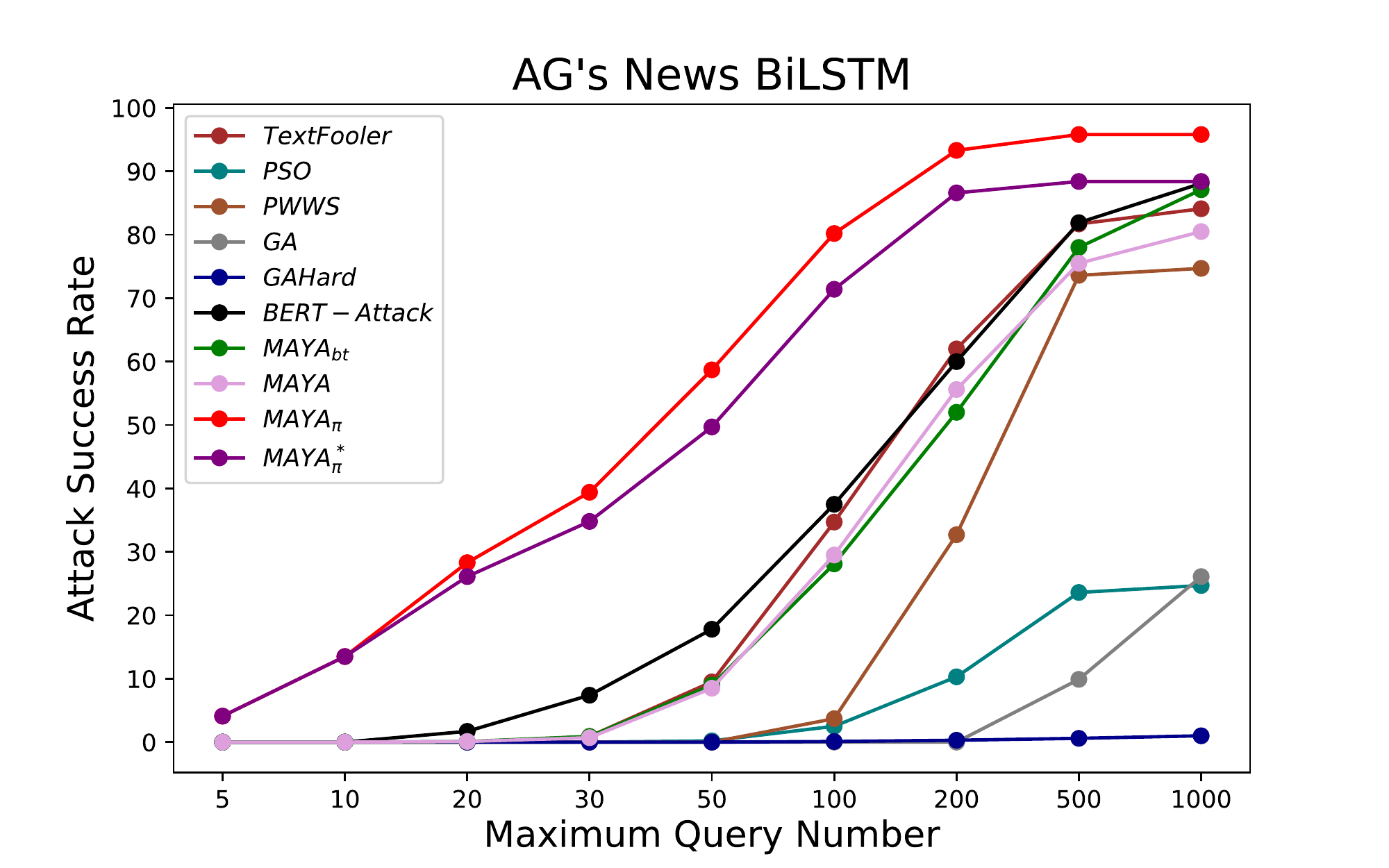}
      \caption{Attack BiLSTM in AG's News}
  \label{fig:minipage6}
\end{minipage}

 \begin{minipage}[b]{0.48\linewidth}
   \includegraphics[height=5.5cm,width=8cm]{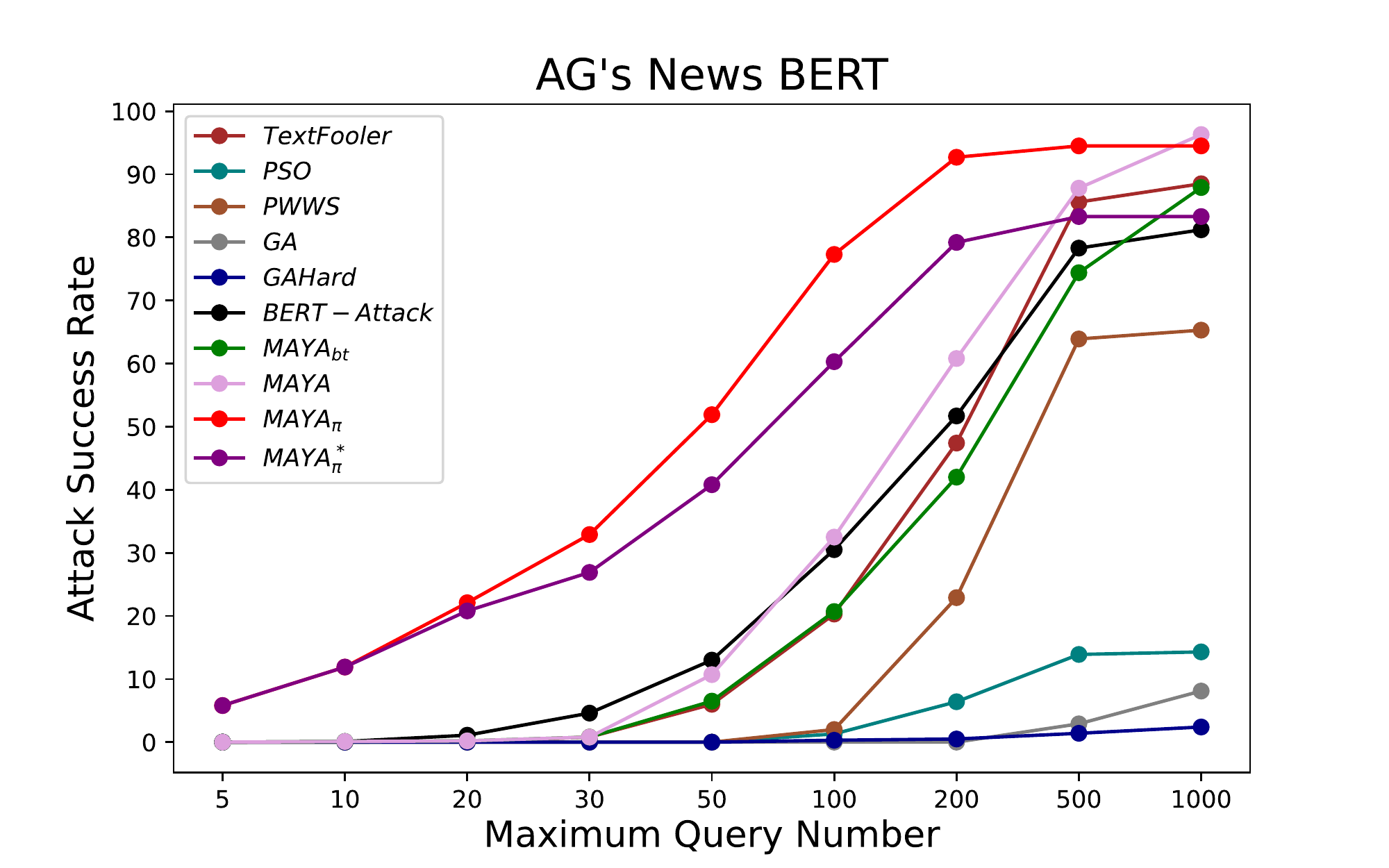}
   \caption{Attack BERT in AG's News}
   \label{fig:minipage7}
 \end{minipage}
 \quad
 \begin{minipage}[b]{0.48\linewidth}
    \includegraphics[height=5.5cm,width=8cm]{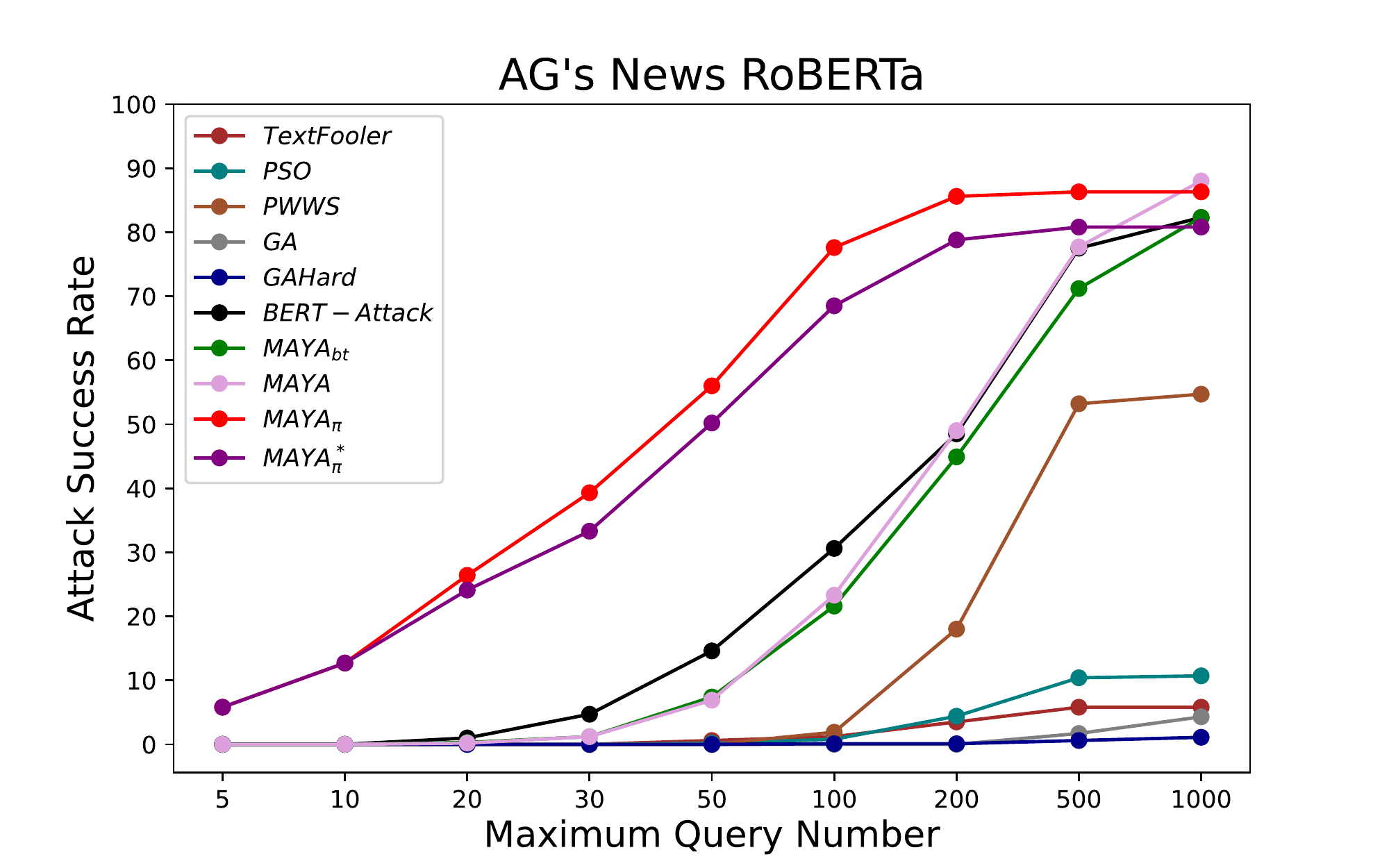}
    \caption{Attack RoBERTa in AG's News}  
  \label{fig:minipage88}
\end{minipage}
\end{figure*}

In this section, we show the results of attack efficiency in three datasets and three victim models. Figure~\ref{fig:minipage1} and Figure~\ref{fig:minipage2} show results of attacking BiLSTM and RoBERTa in SST-2. Figure~\ref{fig:minipage3}-\ref{fig:minipage5} show results of attacking BiLSTM, BERT, and RoBERTa in MNLI. Figure~\ref{fig:minipage6}-\ref{fig:minipage88} show results of attacking BiLSTM, BERT, and RoBERTa in AG's News.

\label{sec:efffigs}

\section{Constituent Selection}
Table \ref{constituentselection} show the selection frequency of all constituent types.
\label{sec:selectfreq}

\label{sec:constituentSelection}

\begin{table}[]
\resizebox{0.48\textwidth}{!}{
\begin{tabular}{
>{\columncolor[HTML]{FFFFFF}}c 
>{\columncolor[HTML]{FFFFFF}}c |
>{\columncolor[HTML]{FFFFFF}}c 
>{\columncolor[HTML]{FFFFFF}}r 
>{\columncolor[HTML]{FFFFFF}}r 
>{\columncolor[HTML]{FFFFFF}}r }
\hline
\cellcolor[HTML]{FFFFFF}                                            & Dataset                           & \multicolumn{4}{c}{\cellcolor[HTML]{FFFFFF}SST-2}                                                                                                                                                                                 \\ \cline{3-6} 
\multirow{-2}{*}{\cellcolor[HTML]{FFFFFF}API}              & Attack Method                     & ASR                                                                        & \multicolumn{1}{c}{\cellcolor[HTML]{FFFFFF}Query} & \multicolumn{1}{c}{\cellcolor[HTML]{FFFFFF}PPL} & \multicolumn{1}{c}{\cellcolor[HTML]{FFFFFF}\%I} \\ \hline
\multicolumn{1}{c|}{\cellcolor[HTML]{FFFFFF}AllenNLP (Score-based)}  & \cellcolor[HTML]{FFFFFF}MAYA$_\pi$   & \multicolumn{1}{r}{\cellcolor[HTML]{FFFFFF}{\color[HTML]{000000} 93.94}} & {\color[HTML]{000000} 21.32}                      & {\color[HTML]{000000} 422}                      & {\color[HTML]{000000} -5.39}                    \\ \hline
\multicolumn{1}{c|}{\cellcolor[HTML]{FFFFFF}Stanza (Decision-based)} & \cellcolor[HTML]{FFFFFF}MAYA$_\pi^*$ & \multicolumn{1}{r}{\cellcolor[HTML]{FFFFFF}{\color[HTML]{000000} 94.16}} & {\color[HTML]{000000} 17.89}                      & {\color[HTML]{000000} 406.36}                   & {\color[HTML]{000000} -4.92}                    \\ \hline
\end{tabular}
}
\caption{\label{attackAPI}	
The results of attacking open-source sentiment analysis models using MAYA$_\pi$.
}

\end{table}

\section{Attack Open-source NLP Frameworks}
In this section, we show that our attack agent can be employed to attack open-source NLP frameworks, including AllenNLP \citep{Gardner2017AllenNLP} and Stanza \citet{qi-etal-2020-stanza}.  We choose the sentiment analysis task and use the attack agent MAYA$\pi$ trained by interacting with BERT trained on SST-2 dataset. Notice that we conduct score-based adversarial attack on AllenNLP sentiment analysis model and decision-based attack on Stanza model according to the outputs of the victim models. 
 Table~\ref{attackAPI} show the results. We observe that our two attack agents can function well in real-world situations in two different attack settings and produce high-quality adversarial samples, showing the potential vulnerability of current NLP systems.

\label{sec:opensourceNLP}

\section{Human Evaluation}
We set up human evaluation to further evaluate the quality of our adversarial samples. Follow \citet{zang-etal-2020-word}, we consider 2 evaluation metrics including validity and naturality. Due to the large number of baseline models, we directly compare our crafted adversarial samples with original samples to evaluate the quality of our adversarial samples. We randomly sample 100 original sentences from SST-2 dataset and 100 adversarial samples crafted by MAYA in SST-2 dataset and mix them. For each sentence, we ask 3 human annotators to do normal sentiment classification task and score this sentence's naturality from 1-5. We use the voting strategy to produce the annotation results of validity for each adversarial sample. Specifically, we respectively measure the human annotators' accuracy on original and adversarial samples and view the difference of accuracy as an indicator of adversarial samples' validity. And we average 3 annotators' naturality scores to get the final results.

From Table~\ref{haman evaluation}, the close gap of accuracy between original and adversarial samples indicates that our adversarial samples maintain high validity. Besides, our adversarial samples also achieve high naturality, which is consistent with automatic evaluation metrics in our main experiments.
\label{sec:humaneval}

\begin{table}[]
\resizebox{0.5\textwidth}{!}{
\begin{tabular}{cc|cc}
\hline
                                             & Victim Model & \multicolumn{2}{c}{BERT}                                                                                  \\ \cline{3-4} 
\multirow{-2}{*}{Dataset}                    & Sample       & Accuracy                                             & Naturality                                          \\ \hline
\multicolumn{1}{c|}{}                        & Original     & \cellcolor[HTML]{FFFFFF}{\color[HTML]{000000} 88.00} & \cellcolor[HTML]{FFFFFF}{\color[HTML]{000000} 4.24} \\
\multicolumn{1}{c|}{\multirow{-2}{*}{SST-2}} & Adversarial  & \cellcolor[HTML]{FFFFFF}{\color[HTML]{000000} 84.00} & \cellcolor[HTML]{FFFFFF}{\color[HTML]{000000} 3.96} \\ \hline
\end{tabular}}
\caption{\label{haman evaluation}	
The results of human evaluation}
\end{table}

\begin{table*}[]
\resizebox{\textwidth}{!}{
\begin{tabular}{cc|rrrr|rrrr|rrrr}
\hline
\rowcolor[HTML]{FFFFFF} 
\cellcolor[HTML]{FFFFFF}                          & Victim Model                   & \multicolumn{4}{c|}{\cellcolor[HTML]{FFFFFF}BiLSTM}                                                                                                                                                                 & \multicolumn{4}{c|}{\cellcolor[HTML]{FFFFFF}BERT}                                                                                                                                                                   & \multicolumn{4}{c}{\cellcolor[HTML]{FFFFFF}RoBERTa}                                                                                                                                                               \\ \cline{3-14} 
\rowcolor[HTML]{FFFFFF} 
\multirow{-2}{*}{\cellcolor[HTML]{FFFFFF}Dataset} & Attack Method                  & \multicolumn{1}{c}{\cellcolor[HTML]{FFFFFF}ASR}            & \multicolumn{1}{c}{\cellcolor[HTML]{FFFFFF}Query} & \multicolumn{1}{c}{\cellcolor[HTML]{FFFFFF}PPL} & \multicolumn{1}{c|}{\cellcolor[HTML]{FFFFFF}\%I} & \multicolumn{1}{c}{\cellcolor[HTML]{FFFFFF}ASR}            & \multicolumn{1}{c}{\cellcolor[HTML]{FFFFFF}Query} & \multicolumn{1}{c}{\cellcolor[HTML]{FFFFFF}PPL} & \multicolumn{1}{c|}{\cellcolor[HTML]{FFFFFF}\%I} & \multicolumn{1}{c}{\cellcolor[HTML]{FFFFFF}ASR}            & \multicolumn{1}{c}{\cellcolor[HTML]{FFFFFF}Query} & \multicolumn{1}{c}{\cellcolor[HTML]{FFFFFF}PPL} & \multicolumn{1}{c}{\cellcolor[HTML]{FFFFFF}\%I} \\ \hline
\rowcolor[HTML]{FFFFFF} 
\cellcolor[HTML]{FFFFFF}                          & SCPN                           & 62.05                                                      & \multicolumn{1}{c}{\cellcolor[HTML]{FFFFFF}-}     & 471.21                                          & \textbf{-10.83}                                  & 52.74                                                      & \multicolumn{1}{c}{\cellcolor[HTML]{FFFFFF}-}     & 467.93                                          & \textbf{-10.26}                                  & 53.25                                                      & \multicolumn{1}{c}{\cellcolor[HTML]{FFFFFF}-}     & 532.45                                          & \textbf{-17.43}                                 \\
\rowcolor[HTML]{FFFFFF} 
\cellcolor[HTML]{FFFFFF}                          & GAhard                         & {\color[HTML]{000000} 91.92}                               & {\color[HTML]{000000} 6584.22}                    & {\color[HTML]{000000} 739.26}                   & {\color[HTML]{000000} 6.03}                      & {\color[HTML]{000000} 81.79}                               & {\color[HTML]{000000} 7460.69}                    & {\color[HTML]{000000} 747.24}                   & {\color[HTML]{000000} 6.81}                      & {\color[HTML]{000000} 75.06}                               & {\color[HTML]{000000} 7589.75}                    & {\color[HTML]{000000} 1179.88}                  & {\color[HTML]{000000} 6.27}                     \\
\rowcolor[HTML]{FFFFFF} 
\multirow{-3}{*}{\cellcolor[HTML]{FFFFFF}SST-2}   & MAYA$_\pi$                     & {\color[HTML]{000000} \textbf{94.21}}                      & {\color[HTML]{000000} \textbf{21.93}}             & {\color[HTML]{000000} \textbf{428.67}}          & {\color[HTML]{000000} -1.59}                     & {\color[HTML]{000000} \textbf{94.88}}                      & {\color[HTML]{000000} \textbf{21.81}}             & {\color[HTML]{000000} \textbf{440.31}}          & {\color[HTML]{000000} -3.81}                     & {\color[HTML]{000000} \textbf{92.92}}                      & {\color[HTML]{000000} \textbf{23.40}}             & {\color[HTML]{000000} \textbf{484.11}}          & {\color[HTML]{000000} -5.37}                    \\ \hline
\rowcolor[HTML]{FFFFFF} 
\cellcolor[HTML]{FFFFFF}                          & \cellcolor[HTML]{FFFFFF}SCPN   & 54.10                                                      & \multicolumn{1}{c}{\cellcolor[HTML]{FFFFFF}-}        & \textbf{570.20}                                 & 4.99                                             & 59.80                                                      & \multicolumn{1}{c}{\cellcolor[HTML]{FFFFFF}-}        & \textbf{465.24}                                 & -1.64                                            & 61.10                                                      & \multicolumn{1}{c}{\cellcolor[HTML]{FFFFFF}-}        & \textbf{446.63}                                 & 1.25                                            \\
\rowcolor[HTML]{FFFFFF} 
\cellcolor[HTML]{FFFFFF}                          & \cellcolor[HTML]{FFFFFF}GAhard & {\color[HTML]{343434} \textbf{74.30}}                      & {\color[HTML]{343434} 6756.56}                    & {\color[HTML]{343434} 1927.06}                  & {\color[HTML]{343434} 9.18}                      & {\color[HTML]{343434} 75.90}                               & {\color[HTML]{343434} 6132.63}                    & {\color[HTML]{343434} 2334.05}                  & {\color[HTML]{343434} 6.17}                      & {\color[HTML]{343434} \textbf{78.50}}                      & {\color[HTML]{343434} 6108.29}                    & {\color[HTML]{343434} 6822.60}                  & {\color[HTML]{343434} 7.91}                     \\
\rowcolor[HTML]{FFFFFF} 
\multirow{-3}{*}{\cellcolor[HTML]{FFFFFF}MNLI}    & MAYA$_\pi$                     & \cellcolor[HTML]{FFFFFF}{\color[HTML]{000000} 72.30}          & {\color[HTML]{000000} \textbf{35.48}}             & {\color[HTML]{000000} 664.00}                   & {\color[HTML]{000000} \textbf{-3.05}}            & \cellcolor[HTML]{FFFFFF}{\color[HTML]{000000} \textbf{76.00}} & {\color[HTML]{000000} \textbf{28.41}}             & {\color[HTML]{000000} 742.28}                   & {\color[HTML]{000000} \textbf{-1.24}}            & {\color[HTML]{000000} 77.60}                               & {\color[HTML]{000000} \textbf{25.91}}             & {\color[HTML]{000000} 693.62}                   & {\color[HTML]{000000} \textbf{-5.84}}           \\ \hline
\rowcolor[HTML]{FFFFFF} 
\cellcolor[HTML]{FFFFFF}                          & SCPN                           & 61.90                                                      & \multicolumn{1}{c}{\cellcolor[HTML]{FFFFFF}-}     & 550.43                                          & \textbf{-28.81}                                  & 52.50                                                      & \multicolumn{1}{c}{\cellcolor[HTML]{FFFFFF}-}     & 649.12                                          & \textbf{-35.33}                                  & 57.50                                                      & \multicolumn{1}{c}{\cellcolor[HTML]{FFFFFF}-}     & 778.08                                          & \textbf{-29.59}                                 \\
\rowcolor[HTML]{FFFFFF} 
\cellcolor[HTML]{FFFFFF}                          & \cellcolor[HTML]{FFFFFF}GAhard & {\color[HTML]{343434} 82.10}                               & {\color[HTML]{343434} 8505.12}                    & {\color[HTML]{343434} 278.51}                   & {\color[HTML]{343434} 3.43}                      & {\color[HTML]{343434} 57.30}                               & {\color[HTML]{343434} 8882.92}                    & {\color[HTML]{343434} 360.20}                   & {\color[HTML]{343434} 4.58}                      & {\color[HTML]{343434} 53.50}                               & {\color[HTML]{343434} 9209.08}                    & {\color[HTML]{343434} 347.19}                   & {\color[HTML]{343434} 3.82}                     \\
\rowcolor[HTML]{FFFFFF} 
\multirow{-3}{*}{\cellcolor[HTML]{FFFFFF}AG's News} & MAYA$_\pi$                     & \cellcolor[HTML]{FFFFFF}{\color[HTML]{000000} \textbf{82.90}} & {\color[HTML]{000000} \textbf{48.75}}             & {\color[HTML]{000000} \textbf{226.21}}          & {\color[HTML]{000000} -8.47}                     & \cellcolor[HTML]{FFFFFF}{\color[HTML]{000000} \textbf{80.60}} & {\color[HTML]{000000} \textbf{64.66}}             & {\color[HTML]{000000} \textbf{246.66}}          & {\color[HTML]{000000} -21.72}                    & \cellcolor[HTML]{FFFFFF}{\color[HTML]{000000} \textbf{76.60}} & {\color[HTML]{000000} \textbf{46.07}}             & {\color[HTML]{000000} \textbf{263.62}}          & {\color[HTML]{000000} -18.66}                   \\ \hline
\end{tabular}

}
\caption{\label{Main experiments results2}	
The results of attack performance and adversarial samples' quality in decision-based attack setting.}

\end{table*}

\section{Case Study}
We select 2 successful adversarial samples crafted by our multi-granularity attack models in AG's News. Note that other baseline attack models all fail in these two samples. We can observe from Table \ref{casestudytable} the strength of our models is twofold. First, from Sample I, our attack models can take different kinds of granularity into consideration, making a bigger search space and crafting more diversified adversarial samples. Second, from Sample II, our attack models can combine different kinds of granularity perturbations to launch a stronger adversarial attack.

\section{Implementation Details}
For our reinforcement learning, we use standard Adam \citep{kingma2014adam} to train our agent and consistently set the learning rate to 2e-5 because our training process is based on data aggregation, meaning that the training data can be abundant. And we set batch size to 16.
 
Due to the limitation of GPU memory and computation resources, we use some tricks to get average batch gradients.     
Given a batch of original sentences list $[S_1, ..., S_i, ..., S_n]$, we input the sentence one by one to our attack agent with all adversarial candidates of original sentences $S_i$ and compute the cross-entropy loss $l_i$ with the golden label from MAYA algorithm. Here, we denote the number of adversarial candidates as $k_i$ and we get weighted loss $L_{i}$ by multiplying $l_i$ with $k_i$:

\begin{equation}
  L_i = l_i \times k_i
\end{equation}
Then we directly perform back propagation to get the gradients for each parameters:
\begin{equation}
g_i = \nabla_{\theta} L_i
\end{equation}
We save the gradients and repeat above operations to accumulate the gradients. Finally, we have:
\begin{equation}
 G = \sum g_i
 \end{equation}
 When reach the batch size, we normalize the gradients and update the parameters.
 \begin{equation}
 G^{'} = \frac{G}{\sum k_i}	 
 \end{equation}


\label{sec:trainingdetails}

\section{Experiment Running Environment}
We conduct all experiments on a server whose major configurations are as follows: (1) CPU: Intel(R) Xeon(R) E5-2620 v4 @ 2.10GHz (2) RAM: 125GB; (3) GPU: RTX1080 , 11GB memory. The operation system is Ubuntu 16.04.7 LTS (GNU/Linux 4.15.0-142-generic x86\_64). We use PyTorch 1.7.1 as the programming framework.

\end{document}